%% file: main.tex
\documentclass[10pt,twocolumn,letterpaper]{article}

\usepackage[pagenumbers]{wacv} %

\input{preamble}
\definecolor{wacvblue}{rgb}{0.21,0.49,0.74}
\usepackage[pagebackref,breaklinks,colorlinks,allcolors=wacvblue]{hyperref}
\usepackage{booktabs}
\usepackage{amsmath}
\usepackage{amssymb}
\usepackage{mathtools}
\usepackage{amsthm}

\input{inwoo}

\newcommand{\ours}{LCBM}
\newcommand{\oursfull}{Locality-aware Concept Bottleneck Model}

\def\wacvPaperID{854} %
\def\confName{WACV}
\def\confYear{2026}

\title{Locality-aware Concept Bottleneck Model}

\author{
Sujin Jeon \quad Hyundo Lee \quad Eungseo Kim \quad Sanghack Lee\thanks{Corresponding authors.}$\:\:$ \quad Byoung-Tak Zhang\footnotemark[1] \\
Seoul National University \\
Seoul, South Korea \\
{\tt\small sn5735@snu.ac.kr} \quad
{\tt\small illhyhl1111@snu.ac.kr} \quad
{\tt\small juniormoo@snu.ac.kr} \\
{\tt\small sanghack@snu.ac.kr} \quad
{\tt\small btzhang@bi.snu.ac.kr}
\and
Inwoo Hwang\footnotemark[1]\\
Columbia University\\
New York, NY\\
{\tt\small inwoo.hwang@columbia.edu}
}

\begin{document}
\maketitle
\begin{abstract}
\input{sections/00-abs}
\end{abstract}

\section{Introduction}
\label{sec:intro}
\input{sections/01-intro}

\section{Related Works}
\label{sec:related-work}
\input{sections/02-related-work}

\section{Method}
\label{sec:method}
\input{sections/03-method}

\section{Experiments}
\label{sec:experiments}

\input{sections/04-experiments}

\section{Limitation}
\label{sec:limitation}
\input{sections/05-limitation}

\section{Conclusion}
\label{sec:conclusion}
\input{sections/06-conclusion}

{
    \small
    \bibliographystyle{ieeenat_fullname}
    \bibliography{ref}
}

\onecolumn
\renewcommand\thesection{\Alph{section}}
\setcounter{section}{0}
\section{Appendix}
\label{sec:appendix}

\input{supp}

\end{document}

%% file: inwoo.tex
\usepackage{amsthm}
\usepackage{amsmath, amssymb, dsfont}
\usepackage{mathrsfs}
\usepackage{graphicx}
\usepackage{wrapfig}
\usepackage{enumitem}

\usepackage{commath}
\usepackage{bbm}
\usepackage{booktabs}
\usepackage{verbatim}
\usepackage{multirow}
\usepackage{makecell}
\usepackage{bbm}
\usepackage{mathtools}
\usepackage{capt-of}
\usepackage{lipsum}
\usepackage{adjustbox}
\usepackage{arydshln}

\usepackage{thmtools,thm-restate}
\usepackage{apxproof}

\usepackage[capitalize]{cleveref}
\Crefname{figure}{Fig.}{Figs.}
\Crefname{section}{Sec.}{Sec.}
\Crefname{equation}{Eq.}{Eqs.}
\Crefname{proposition}{Prop.}{Props.}
\Crefname{prop}{Prop.}{Props.}
\Crefname{theorem}{Thm.}{Thms}
\Crefname{lemma}{Lemma}{Lemmas}
\Crefname{definition}{Def.}{Defs}
\Crefname{algorithm}{Alg.}{Algs}
\Crefname{remark}{Remark}{Remarks}
\Crefname{example}{Example}{Examples}
\Crefname{corollary}{Corollary}{Corollary}
\Crefname{Appendix}{App.}{Apps.}
\Crefname{assumption}{Assumption}{Assumptions}

\usepackage{pifont}
\usepackage{pdfrender}
\usepackage{colortbl}
\definecolor{forestgreen}{rgb}{0.13, 0.55, 0.13}

\usepackage{algorithm}
\usepackage{algorithmic}
\usepackage{fancyhdr}
\usepackage{color, colortbl}
\definecolor{Gray}{gray}{0.9}
\definecolor{lightgray}{rgb}{0.83, 0.83, 0.83}

\newcommand{\stdv}[1]{\scriptsize$\pm$#1}

\newcommand{\paragrapht}[1]{\paragraph{#1}}

\def\*#1{\mathbf{#1}}

\usepackage{amsmath,amsfonts,bm}

\def\eqref#1{equation~\ref{#1}}

\def\1{\bm{1}}

\DeclareMathAlphabet{\mathsfit}{\encodingdefault}{\sfdefault}{m}{sl}
\SetMathAlphabet{\mathsfit}{bold}{\encodingdefault}{\sfdefault}{bx}{n}

\def\gC{{\mathcal{C}}}

\def\gL{{\mathcal{L}}}

%% file: sections/00-abs.tex
Concept bottleneck models (CBMs) are inherently interpretable models that make predictions based on human-understandable visual cues, referred to as \textit{concepts}.
As obtaining dense concept annotations with human labeling is demanding and costly, recent approaches utilize foundation models to determine the concepts existing in the images.
However, such label-free CBMs often fail to localize concepts in relevant regions, attending to visually unrelated regions when predicting concept presence.
To this end, we propose a framework, coined \oursfull{} (\ours), 
which utilizes rich information from foundation models and adopts prototype learning to 
ensure accurate spatial localization of the concepts.
Specifically, we assign one prototype to each concept, promoted to represent a prototypical image feature of that concept. 
These prototypes are learned by encouraging them to encode similar local regions, leveraging foundation models to assure the relevance of each prototype to its associated concept.
Then we use the prototypes to facilitate the learning process of identifying the proper local region from which each concept should be predicted. 
Experimental results demonstrate that \ours{} effectively identifies present concepts in the images and exhibits improved
localization while maintaining comparable classification performance.

%% file: sections/01-intro.tex
The interpretability of deep neural networks has become an increasingly important issue as the rapid advancements in AI make them closely integrated into our daily lives. 
This is especially critical in fields where the reliability of models is paramount, e.g., healthcare and social sciences.
Since explaining the decision-making process of a black-box model is often challenging, 
several lines of research focus on building inherently interpretable models whose decision-making is naturally understandable to humans.

\begin{figure}[t]
\begin{center}
\includegraphics[width=\linewidth]{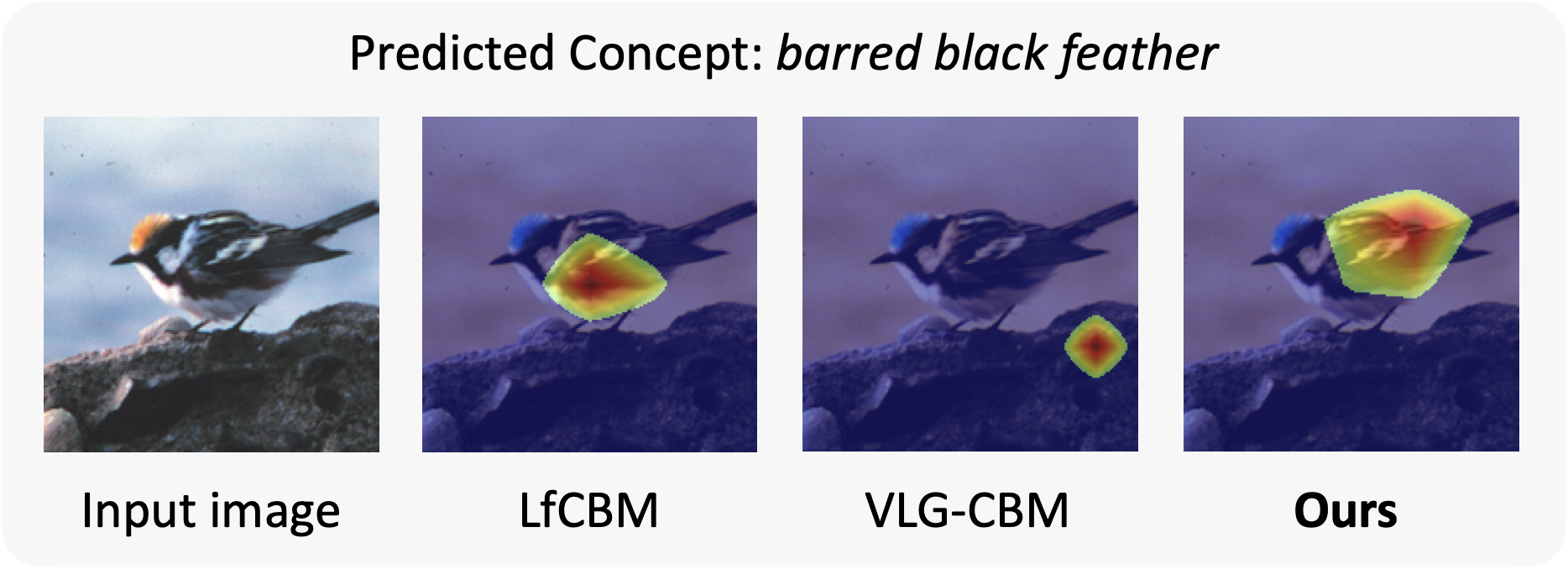}
\caption{Concept localization in label-free CBMs. The leftmost image is the input to the models, while the others show the saliency maps for the concept ``barred black feather''. 
The activated area in our model is well-aligned with the concept, in contrast to prior label-free CBMs focusing on irrelevant regions.
}
\label{fig:figure-1}
\end{center}
\end{figure}

Concept bottleneck model (CBM) \citep{cbm,cem, coopcbm, ecbm, probcbm} is a representative inherently-interpretable architecture where a \textit{concept} refers to human-recognizable visual cues. 
CBMs first predict the concepts existing in the image (e.g.,``red color'' and ``round shape'' in the image of an apple) and make a final prediction on the class label through the linear combination of these concepts. 
Thus, they offer explanations for model decisions through the concept bottleneck.
However, their reliance on dense concept annotations
limits their scalability and practicality. This has led to the emergence of label-free CBMs \citep{labelfree,labo,concise,faithful,rescbm, vlgcbm}, which utilize foundation models such as large language models (LLMs) and vision-language models (VLMs) to determine and estimate the presence of concepts.

Despite their promises, existing label-free CBMs often fail to properly localize the concepts in the corresponding regions of an image.
For example, they attend to irrelevant spatial regions for the prediction of the concept ``barred black feather'', as shown in \Cref{fig:figure-1}. 
Such misalignment raises critical concerns about the reliability of their explanations.

To this end, we propose a framework, coined \oursfull{} (\ours). 
In the \ours{} framework, the relationship between the concepts and local regions in the image is learned. 
To achieve this, we incorporate prototype learning which additionally leverages the rich prior knowledge of CLIP \citep{clip}. Specifically, each prototype is assigned to a single concept, encouraged to represent prototypical image features of that concept.
The learning process of prototypes includes an auxiliary classification task, facilitating the prototypes to encode features of visually similar local regions.
Meanwhile, CLIP similarity scores are used to ensure the alignment of prototypes to the corresponding concepts.
Finally, the similarity between a local region and a prototype determines the relevance of the local region to the concept associated with the prototype.
With this strategy, \ours{} can effectively refer to related regions when predicting concepts, making the explanation more reliable.

To demonstrate the effectiveness of \ours{}, we evaluated the quality of concept localizations using two protocols. First, we adopted GradCAM \citep{gradcam}
to identify the activated regions for the concept predictions. We then compared these activated regions against ground-truth points or bounding boxes of present concepts. Second, we tested whether the model truly relied on the corresponding region when predicting present concepts, by removing the concepts from the image and observing the decrease in concept presence scores. Our experiments on these protocols demonstrate that the concept predictions of \ours{} are properly localized in the image. 

We also evaluated the effectiveness of explanations (i.e., whether the model utilizes existing concepts when making the decision), which is a key requirement in CBMs. For this, we calculated the precision of concepts which significantly contributed to the decision using a Multimodal Large Language Model. The results showcase that \ours{} has a strong capability in explaining the model decision with relevant concepts, thus providing effective explanations.

Our main contributions are summarized as:
\begin{itemize}
    \item We addressed the issue of concept localization in label-free CBMs, which has not yet been explored in existing research.
    \item We introduce \oursfull{}, a novel framework promoting concept predictions to be more localized in the corresponding regions by incorporating a prototype learning strategy.
    \item We experimentally demonstrated that the concept predictions of our method are more properly localized in the image, while also showcasing a strong capability in explaining the model decision with relevant concepts.
\end{itemize}

%% file: sections/02-related-work.tex
\paragrapht{Concept Bottleneck Models (CBMs).}
CBM \citep{cbm, probcbm,cem,coopcbm,ecbm} is an inherently interpretable framework that makes predictions based on the \textit{concepts}, i.e., canonical visual cues composing the objects and the scene.
CBMs first predict which concepts exist in the image and then make a final prediction based solely on these predicted concepts, enabling model decisions to be explainable in terms of concept presence.

\paragrapht{Label-free CBMs.}
A significant drawback of conventional CBMs is that they rely on dense concept annotations, requiring extensive human labor.
To address this, recent label-free CBMs \citep{pcbm,labelfree,labo,concise,faithful,rescbm, vlgcbm} utilize foundation models such as CLIP \citep{clip} to determine concept presence without human annotation,
extending CBMs to diverse domains and data scales.
Recently, VLG-CBM \citep{vlgcbm} utilizes an open-domain object detection module to incorporate supervision on concept predictions with bounding-box annotations.
However, as shown in \Cref{fig:figure-1}, ensuring accurate concept localization in label-free CBMs remains an unresolved challenge, raising concerns about the reliability of the explanations these models provide. 

\paragrapht{Credibility in CBMs.}
Several approaches have focused on credibility in CBMs, such as the faithfulness \citep{faithful} and trustworthiness \citep{protocbm} of explanations. 
ProtoCBM \citep{protocbm} explored the issue of concept trustworthiness, emphasizing that conventional CBMs often fail to reference corresponding regions when predicting concepts, a problem also noted in previous works \citep{locality, intended}. This concern aligns closely with the problem we address. However, our work focuses on the label-free setting, 
which is a more challenging scenario
as no explicit guidance exists within the concept bottleneck.
Recently, SALF-CBM \citep{salfcbm} 
aims to resolve the issue of concept localization,
but
it cannot determine which concept is most likely to be present in a specific region. In comparison, by introducing learned prototypes, \ours{} affords a richer, more interpretable representation, since it allows us to examine precisely how each image patch relates to individual concepts. Additionally, to the best of our knowledge, this is the first work to perform a quantitative analysis of concept-level localization.

\paragrapht{Prototype-based explanations.}
Prototype-based explanation methods \citep{protopnet, protoconcepts, pipnet} represent another line of research that shares a common feature with CBMs: inherent interpretability. These methods explain model decisions by associating local regions in an image with similar image patches. They are analogous to our work as they also use prototypes as representatives of similar local regions. However, unlike our approach, the prototypes in these methods are not explicitly represented by text, requiring humans to interpret them post hoc. Additionally, these prototypes may not always be as interpretable as those explicitly aligned with textual descriptions.

\begin{figure*}[t]
\vskip 0.1in
\begin{center}
\centerline{\includegraphics[width=0.9\textwidth]{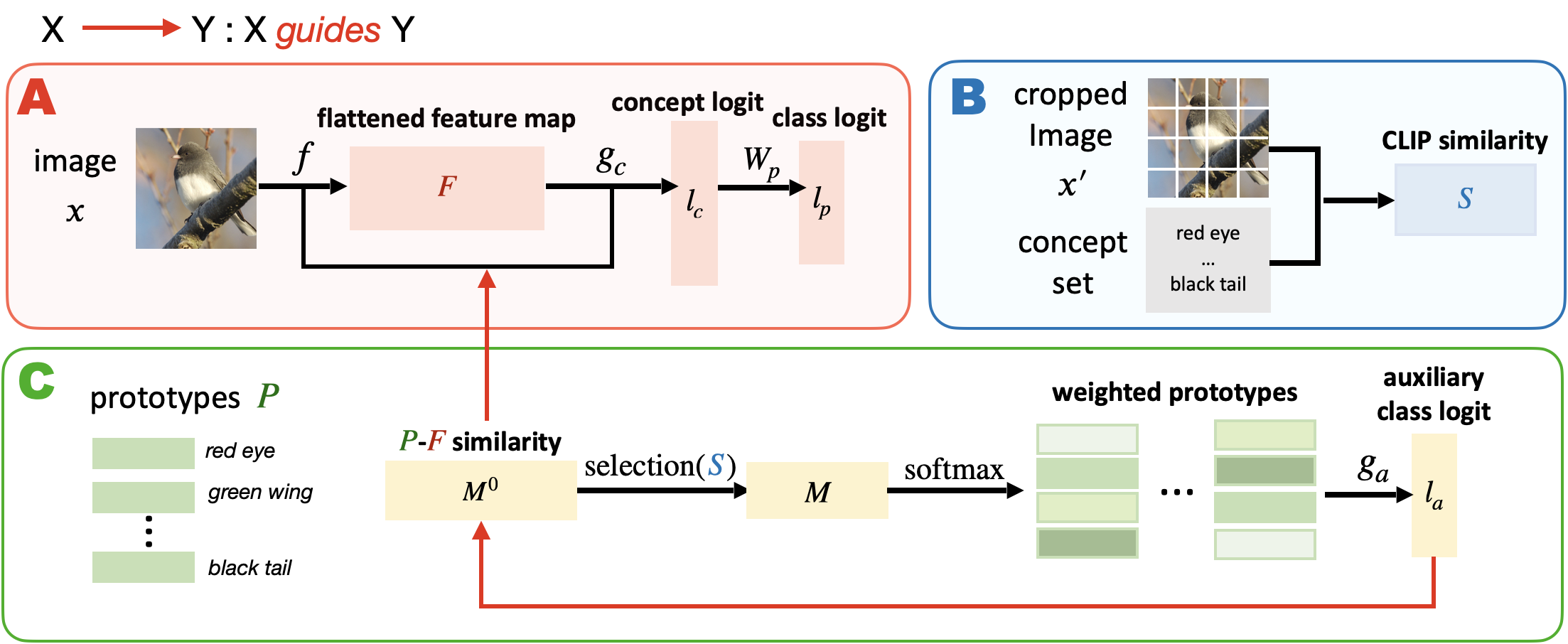}}
\caption{Overview of our method. (A) Given an input image $x$, we first extract features $F$ and obtain concept logit $l_c$ and class label logit $l_p$ by sequentially applying linear transformations. 
(B) In parallel, we crop the image into $H\times W$ patches
and yield concept-patch similarity matrix using CLIP.
(C) We introduce prototypes $P$, where each prototype is assigned to a concept. $P$ and $F$ are dot-producted to obtain prototype-patch similarity matrix $M^0$. $M^0$ is compressed to $M$ by gathering prototypes of top-$K_1$ concepts for patch $(h,w)$ from $S$. Then we apply softmax to $M$ along the $K_1$ dimension, using the resulting values to weight the $K_1$ prototypes for each local region $(h,w)$. The weighted prototypes are averaged, and a linear transformation $g_a$ is applied to produce the auxiliary classification logits $l_a$. 
(Red Arrows) $M^0$ is guided by the auxiliary classification loss using $l_a$. Finally, backbone $f$ and linear transformation layer $g_c$ is guided by $M^0$, effectively localizing concept predictions in corresponding regions.
}
\label{fig:overall-architecture}
\end{center}
\vskip -0.1in
\end{figure*}

%% file: sections/03-method.tex
In this section, we present our framework whose prediction is based on concepts, each derived from highly relevant regions.
First, we describe the curation of a concept set using LLM (\Cref{sec:method-concept}).
Then, we detail each module of our framework, followed by a discussion of the losses that align these modules (\Cref{sec:method-arch,sec:method-loss}).
The overall architecture of our method is illustrated in \Cref{fig:overall-architecture}.

\subsection{Concept Set Generation}
\label{sec:method-concept}
We employ GPT-4o  \citep{gpt4} to curate a set of concepts $\gC=\{c_1, c_2, \cdots c_K\}$, similar to previous label-free CBMs \citep{labo,labelfree, vlgcbm, concise}. 
Specifically, we generate a compositional concept set where each concept consists of canonical attributes and parts that are useful to predict the class label.
For instance, in a dataset where the classes are different bird species, the examples of concepts can be ``black beak'' or ``rounded head''. Details on the concept set generation are provided in Appendix A.

\subsection{Overall Architecture}
\label{sec:method-arch}
\subsubsection{Concept Bottleneck Architecture}
\label{sec:method-arch-a}
We employ a concept bottleneck architecture that first predicts concept presences and then makes a final prediction on class labels based on concepts, as depicted in \Cref{fig:overall-architecture}-(A).

For each sample $(x, y)$, where $x$
is the input image and $y$ is the corresponding label, we first extract features of $x$ with a backbone network $f$. 
This yields a flattened feature map $F \in \mathbb{R}^{HW \times D}$, where $H$ and $W$ are the spatial dimensions of feature map before flattening, and $D$ is the feature dimension.
We apply average pooling along the $HW$ dimension of $F$ and 
apply the linear layer $g_c$ to obtain the concept logit $l_c$.
Through the final linear layer with weight $W_p$, we have the class logit $l_p = W^\top_p l_c$. 
We use standard cross-entropy loss for the classification loss:
\begin{align}\label{eq:loss-classification}
\gL_{class}= \gL_{CE}(\hat{y}, y),
\end{align}
where $\hat{y}$ is the label prediction of the model (i.e., softmax over class logit $l_p$).

Building upon this architecture, we aim to improve the localization of concept predictions. To achieve this, we employ a prototype learning integrated with CLIP scores, which will be described in the following sections.

\subsubsection{Patch-wise Concept Similarity}
\label{sec:method-arch-b}
As illustrated in \Cref{fig:overall-architecture}-(B),
we crop the image into $HW$ patches and extract the CLIP features for each patch, resulting in $F_I \in \mathbb{R}^{HW \times D_c}$, where $D_c$ is the CLIP feature dimension.
Similarly, we compute the CLIP features for the concept set $\mathcal{C}$, yielding $F_T \in \mathbb{R}^{K \times D_c}$. 
We then obtain the CLIP score matrix $S\in \mathbb{R}^{HW\times K}$ where each element $S_{hw,k}$ is the cosine similarity of $\left(F_{I}\right)_{hw}$ and $\left(F_T\right)_{k}$. %
In our framework, each $(h,w)$ image patch corresponds to a specific local region represented by $F_{hw}$.

\subsubsection{Prototype Learning}
\label{sec:method-arch-c}
We employ prototypes $P=\{p_1, \cdots, p_K\}$, where each prototype $p_k\in\mathbb{R}^D$ corresponds to a concept $c_k$.  
These prototypes 
are encouraged to serve as prototypical image features of the associated concept.

To facilitate the prototypes to first encode the features of similar local regions, 
we adopt an auxiliary classification task. Logits for this task are derived from weighted prototypes for each local region, with weights based on the similarity between the prototypes and the local region.
Thus, these similarities are learned through the auxiliary classification task, as it ensures high classification accuracy when the weights of prototypes crucial to the local region are high.
However, without proper guidance, a prototype may represent local regions that are irrelevant to the concept it was initially assigned to. %

To address this, we first select a meaningful subset of prototypes using CLIP scores $S$. This provides guidance on which prototypes should be considered as potentially crucial for a given local region.
With this guidance, the local region and the prototypes for concepts that are highly relevant to that region are promoted to have high weights (i.e., similarities).

The detailed procedure is outlined as follows. First, we initialize $K$ prototypes each having the same feature dimension $D$ as feature map $F$. Then we compute cosine similarity of $F$ and $P$, resulting in a similarity score matrix $M^0\in \mathbb{R}^{HW\times K}$. Here, we utilize CLIP scores $S$ to get only top-$K_1$ elements along the $K$ concepts. Specifically, for each $(h,w)$ patch, we obtain top-$K_1$ concepts with highest values in $S_{hw}$. 
This enables the prototypes to focus on the local regions that are meaningful for the associated concepts. 
We gather $K_1$ columns of $M^0$, which correspond to top-$K_1$ concepts, obtaining $M\in \mathbb{R}^{HW\times K_1}$. 

Then we use $M$ to weight $K_1$ prototypes, where each prototype represents each of the top-$K_1$ concepts. In detail, for each local region $(h,w)$, we apply softmax function to $M_{hw}$ as the weights for $K_1$ prototypes and multiply them with those prototypes. After summation of weighted prototypes, we average them over $HW$ local regions and map it to auxiliary classification logit $l_a$ by applying linear transformation layer $g_a$.

\subsection{Loss}
\label{sec:method-loss}
We now introduce the losses designed to encourage the model to focus on the relevant regions when predicting concepts and to effectively identify which concepts exist in the image. First, we use standard cross-entropy loss for auxiliary classification loss $\gL_{aux}$ as follows:
\begin{align}\label{eq:loss-auxiliary}
\gL_{aux}= \gL_{CE}(\hat{y}_{aux}, y),
\end{align}
where $\hat{y}_{aux}$ is the auxiliary label prediction of the model. (i.e., softmax over auxiliary class logit $l_a$)

We then introduce a loss that guides $f$ and $g_c$ through $M^0$. With $M_{hw}$ representing the relevance of $K_1$ selected prototypes for each $(h,w)$ local region, we assume that top-$K_2$ prototypes are more significant for each region. Therefore, we assign negative values to the elements in $M^0_{hw}$ that are not in $\mathrm{argtop}K_2(M_{hw})$, ensuring that they become zero when applying softmax. This results in $M^{0^{\prime}}$.

In parallel, we compute the influence value of each local region $(h,w)$ on the prediction of the concept $c_k$ as:
\begin{equation}
V_{k,hw}=\sum_{d=1}^{D}F_{hw,d}\frac{1}{HW}\sum_{h'=1}^{H}\sum_{w'=1}^{W}\left[\frac{\partial{l_c}}{\partial{F}}\right]_{k,h'w',d}.
\end{equation}
A larger $V_{k,hw}$ indicates that the local region $(h,w)$ has a higher relevance on the prediction of concept $c_k$. Since our goal is to ensure that concept predictions are localized in highly relevant regions, we minimize the distribution distance between $V$ and $M^{0^{\prime}}$. This is formulated as follows:  
\begin{align}\label{eq:loss-local}
\gL_{local}=\sum_{k=1}^K D_{KL}\left(V_k \parallel M^{0^{\prime}}_{:, k}\right),
\end{align}
where $D_{KL}$ is the Kullback–Leibler divergence after applying softmax function to both.

The final loss is $\gL_{total} = \gL_{class} + \alpha\gL_{local} + \beta\gL_{aux}$, where $\alpha$ and $\beta$ are balancing coefficients. The coefficients we used are detailed in Appendix C.

%% file: sections/04-experiments.tex
\subsection{Setup}
\paragrapht{Datasets.}
We used four datasets to evaluate the performance and explanability of models.
\textbf{CUB-200-2011} \citep{cub} 
contains 11788 images across 200 bird species, with 5994 images for training. 
\textbf{ImageNet-animal} is a subset of ImageNet \citep{imagenet} dataset containing 398 animal categories.
It consists of 510530 training images and 19900 validation images.  
\textbf{Stanford Cars} \citep{cars} consists of 16185 images in total, with 8144 training images across 196 classes of various car models. 
Since the images in these datasets can be explained as a combination of different parts, it is crucial that the concepts are properly localized in the corresponding parts to ensure reliable explanation.

Finally, we consider 
\textbf{Fitzpatrick17k} \citep{fitz}, a skin disease diagnosis dataset 
with 16577 images in total. 
We randomly split the dataset into 9896 train images and 6681 validation images. 
The images involve varying lesion locations, and thus,
the model should identify the lesion's location and predict concepts based on it for explanation reliability.

\paragrapht{Baselines.}
We compared \ours{} with LfCBM \citep{labelfree} and LaBo \citep{labo}, 
the representative label-free CBMs. 
We also include VLG-CBM \citep{vlgcbm}, a recent method that improves the alignment between predicted concepts and the image by utilizing bounding-box annotations with an open-domain object detection module. 

Although LaBo does not train concept prediction layer and was therefore excluded from localization evaluations, we included it as a baseline to compare with a model without backbone.
Implementation details are provided in Appendix C.
Additional experimental results are provided in Appendix D.

\subsection{Evaluation Protocol}
\label{sec:evaluation-protocol}
\paragraph{Precision evaluation} 
To evaluate whether the concepts highly attributing to the model decision truly exist in the image, we measured the precision of the concepts having high contributions (i.e., product of the concept presence score and corresponding weight to the class).
Specifically, we selected the top-$k$ concepts in terms of their contributions and queried GPT-4o \citep{gpt4} to determine their presence.
Finally, we computed the precision as the ratio of the number of concepts deemed valid by GPT-4o to the total number of top-$k$ concepts for each image.
Details are provided in Appendix B.1.

\paragraph{Localization evaluation} 
To evaluate the model's localization ability of the concepts, we generated GradCAM \citep{gradcam} maps for concepts from the last convolutional block of the backbone network. These maps were generated for concepts determined by GPT-4o to exist in the image, following the precision evaluation protocol. By applying a threshold to the GradCAM maps, we extracted the highest activated areas corresponding to each concept.

For this evaluation, we sampled a subset of images from each dataset and annotated points and bounding boxes indicating where the concepts exist. Since the concepts consist of attribute-part combinations (e.g., ``striped blue wing''), we annotated the location of each part rather than each individual concept, except for Fitzpatrick17k. For Fitzpatrick17k, we annotated the location of each lesion. Using these annotations, we evaluated the model's performance on four metrics: Inclusion, mIoU, REP, and Deletion.

\begin{itemize}[leftmargin=*]
    \item \textit{Inclusion}: This metric calculates the ratio of concepts whose activated area contains the corresponding ground-truth part point. For instance, for the concept ``barred red tail'', the activated area should include the point indicating the location of the part ``tail''.
    \item \textit{mIoU}: We computed the mean Intersection over Union (IoU) between the activated area and the annotated bounding box.
    \item \textit{REP}: This metric measures the ratio of the intersection area between the activated region and the annotated bounding box to the area of the bounding box itself.
    \item \textit{Deletion}: To evaluate whether the model inherently focuses on the corresponding region when predicting concepts, we visually deleted the region associated with each concept and analyzed the resulting changes in concept scores. Specifically, we zeroed out the pixel values inside the bounding box corresponding to each present concept.
    We then calculated the ratio and difference in concept scores before and after deletion, with sigmoid applied to each score. If the model does not attend to the corresponding region when predicting a concept, the ratio will remain high, and the difference will be low.
\end{itemize}

Inclusion, mIoU, and REP assess concept localization ability in the context of post-hoc analysis, while Deletion evaluates it in the context of inherent analysis. Therefore, the two protocols complement each other.
Details of the localization evaluation are provided in Appendix B.2.

\begin{table*}[ht]
\centering
\caption{Accuracy and precision evaluation results.}
\label{table:main}
\begin{adjustbox}{width=0.8\textwidth}%
\begin{tabular}{lcccccccc}
\toprule
& \multicolumn{2}{c|}{CUB-200-2011} & \multicolumn{2}{c|}{ImageNet-animal} & \multicolumn{2}{c|}{Stanford Cars} & \multicolumn{2}{c|}{Fitzpatrick17k} \\ \cmidrule(lr){2-3} \cmidrule(lr){4-5} \cmidrule(lr){6-7} \cmidrule(lr){8-9}
Model & Accuracy & Precision & Accuracy & Precision & Accuracy & Precision & Accuracy & Precision\\ \midrule
LfCBM \citep{labelfree} & 73.23\stdv{0.2} & 44.73\stdv{0.5} & \underline{83.64}\stdv{0.1} & 57.79\stdv{0.7} & 70.16\stdv{0.4} & 59.34\stdv{1.1} & 47.71 \stdv{0.9} & 46.83 \stdv{1.8} \\
LaBo \citep{labo} & 73.14\stdv{0.1} & \underline{72.81}\stdv{0.1} & 79.76\stdv{0.0} & 64.22\stdv{0.6} & 71.34\stdv{0.8} & 64.92\stdv{2.4} & 38.51\stdv{0.6} & 48.71\stdv{1.8} \\
VLG-CBM \citep{vlgcbm} & \textbf{76.54}\stdv{0.1} & 68.19\stdv{0.7} & \textbf{84.24}\stdv{0.3} & \underline{75.72}\stdv{1.1} & \textbf{81.32}\stdv{0.0} & \textbf{68.93}\stdv{0.5} & \textbf{51.66}\stdv{0.1} & \underline{50.49}\stdv{0.9} \\
\ours{} (Ours) & \underline{75.24}\stdv{0.4} & \textbf{76.59}\stdv{0.2} & 83.59\stdv{0.1} & \textbf{76.05}\stdv{0.4} & \underline{79.30}\stdv{0.2} & \underline{65.47}\stdv{0.7} & \underline{50.21}\stdv{0.2} & \textbf{53.29}\stdv{1.5} \\ 
\bottomrule
\end{tabular}
\end{adjustbox}
\vskip -0.05in
\end{table*}

\begin{table*}[ht]
\centering
\caption{Localization evaluation results.}
\label{table:localization}
\begin{adjustbox}{width=\textwidth}%
\begin{tabular}{lcccccccccccc}
\toprule
& \multicolumn{3}{c}{CUB-200-2011} & \multicolumn{3}{c}{ImageNet-animal} & \multicolumn{3}{c}{Stanford Cars} & \multicolumn{3}{c}{Fitzpatrick17k} \\ \cmidrule(lr){2-4} \cmidrule(lr){5-7} \cmidrule(lr){8-10} \cmidrule(lr){11-13}
Model & Inclusion & mIoU & REP & Inclusion & mIoU & REP & Inclusion & mIoU & REP & Inclusion & mIoU & REP  \\ \midrule
LfCBM & 65.34\stdv{2.1} & 28.33\stdv{0.5} & 37.62\stdv{0.7} & 56.94\stdv{2.8} & 26.29\stdv{0.6} & 40.77\stdv{1.0} & 38.30\stdv{0.2} & 18.16\stdv{0.4} & 26.80\stdv{0.5} & \textbf{86.58}\stdv{1.0} & 35.32\stdv{2.1} & 45.95\stdv{3.0}\\
VLG-CBM & 53.81\stdv{0.7} & 22.71\stdv{0.2} & 30.63\stdv{0.2} & 54.84\stdv{3.1} & 23.78\stdv{0.7} & 40.48\stdv{1.8} & 20.87\stdv{0.4} & 11.65\stdv{0.1} & 16.13\stdv{0.3} & 64.77\stdv{0.1} & 22.12\stdv{0.4} & 26.45\stdv{0.5}\\
\ours{} (Ours) & \textbf{72.44}\stdv{1.7} & \textbf{29.74}\stdv{0.2} & \textbf{51.17}\stdv{0.9} & \textbf{82.44}\stdv{0.1} & \textbf{28.61}\stdv{0.4} & \textbf{70.07}\stdv{0.6} & \textbf{71.58}\stdv{0.1} & \textbf{30.40}\stdv{0.2} & \textbf{53.19}\stdv{0.2} & 84.55\stdv{0.7} & \textbf{36.86}\stdv{0.4} & \textbf{49.88}\stdv{0.6}\\
\bottomrule
\end{tabular}
\end{adjustbox}
\vskip -0.1in
\end{table*}

\subsection{Results}
\paragraph{Classification accuracy}
The `Accuracy' columns in \Cref{table:main} show the classification accuracy for each dataset.
\ours{} achieved comparable accuracies to the baselines, demonstrating its ability to maintain performance. 

\paragraph{Main results}
\Cref{table:main} presents the precision evaluation results, and \Cref{table:localization} shows the localization evaluation results.
As \Cref{table:main} demonstrates, \ours{} attains higher precision than all baselines in three datasets. This underscores the strength of our framework, achieving robust explanation capabilities by the utilization of concepts truly exist in the image when making the decision. 

\Cref{table:main,table:localization} show a trade-off between explanation capability and the localization of concepts in baselines. 
Specifically, although VLG-CBM achieves high precision, it exhibits relatively low localization scores. In case of LfCBM, it achieves higher localization scores than VLG-CBM but struggles to effectively utilize present concepts in the image, as demonstrated by the precision results. In contrast, \ours{} achieves effective alignment between utilized concepts and the image while simultaneously ensuring its reliability through accurately localized concept predictions. 

For Stanford Cars, \ours{} has a lower precision score than VLG-CBM. 
However, \Cref{table:localization} demonstrates that \ours{} achieves significantly superior localization ability compared to all baselines in Stanford Cars, highlighting its ability to provide reliable explanations while maintaining reasonably strong overall performance.

\begin{table*}[h]
\centering
\caption{Deletion results.}
\label{table:deletion}
\begin{adjustbox}{width=0.95\textwidth}%
\begin{tabular}{lcccccccc}
\toprule
& \multicolumn{2}{c}{CUB-200-2011} & \multicolumn{2}{c}{ImageNet-animal} & \multicolumn{2}{c}{Stanford Cars} & \multicolumn{2}{c}{Fitzpatrick17k} \\ \cmidrule(lr){2-3} \cmidrule(lr){4-5} \cmidrule(lr){6-7} \cmidrule(lr){8-9}
Model & Ratio$\downarrow$ & Difference$\uparrow$ & Ratio$\downarrow$ & Difference$\uparrow$ & Ratio$\downarrow$ & Difference$\uparrow$ & Ratio$\downarrow$ & Difference$\uparrow$ \\ \midrule 
LfCBM & 87.17\stdv{0.2} & 0.1103\stdv{0.002} & 93.59\stdv{0.5} & 0.0587\stdv{0.004} & 95.27\stdv{0.2} & 0.0374\stdv{0.001} & 80.86\stdv{0.3} & 0.1635\stdv{0.003}\\
VLG-CBM & 90.36\stdv{0.2} & 0.0863\stdv{0.002} & 98.67\stdv{0.1} & 0.0131\stdv{0.001} & 91.28\stdv{0.2} & 0.0728\stdv{0.002} & 88.72\stdv{0.5} & 0.1039\stdv{0.005}\\
\ours{} (Ours) & \textbf{79.25}\stdv{3.5} &  \textbf{0.1772}\stdv{0.028} &  \textbf{88.38}\stdv{1.1} &  \textbf{0.0984}\stdv{0.009} &  \textbf{82.14}\stdv{0.5} &  \textbf{0.1564}\stdv{0.004} &  \textbf{65.80}\stdv{0.8} &  \textbf{0.2726}\stdv{0.004} \\ 
\bottomrule
\end{tabular}
\end{adjustbox}
\vskip -0.05in
\end{table*}

\paragraph{Results on deletion}
\Cref{table:deletion} presents the results for the deletion. The results show that \ours{} achieves a lower ratio and a higher difference compared to the baselines across all datasets, indicating that \ours{} inherently focuses more on the corresponding regions when predicting present concepts. 

\subsection{Ablation Study}
\begin{table*}[h]
\centering
\caption{Ablation study on CUB-200-2011.}
\label{table:ablation}
\begin{adjustbox}{width=0.65\textwidth}%
\begin{tabular}{cccccccc}
\toprule
& & & & & \multicolumn{3}{c}{Localization evaluation} \\  \cmidrule(lr){6-8}
Patch size & $K_1$ & $K_2$ & Accuracy & Precision & Inclusion & mIoU & REP \\ \midrule 
32 & 5 & 2 & 73.57\stdv{1.0} & 77.57\stdv{0.4} & 71.50\stdv{1.5} & 27.98\stdv{0.9}& 43.45\stdv{1.2} \\
32 & 5 & 3 & 74.40\stdv{0.2} & 77.64\stdv{0.2} & 71.32\stdv{0.8} & 27.93\stdv{0.1} & 43.22\stdv{0.5} \\
32 & 7 & 2 & 74.47\stdv{0.6} & 77.38\stdv{0.6} & 70.90\stdv{0.2} & 27.75\stdv{0.8} & 43.65\stdv{1.3} \\
32 & 7 & 3 & 74.62\stdv{0.5} & 77.28\stdv{1.8} & 71.72\stdv{0.2} & 27.98\stdv{1.0} & 44.07\stdv{1.0} \\ \midrule
56 & 5 & 2 & 75.39\stdv{0.3} & 76.47\stdv{0.3} & 69.02\stdv{1.9} & 28.95\stdv{1.1} & 48.23\stdv{2.0} \\
56 & 5 & 3 & 75.14\stdv{0.3} & 77.13\stdv{0.6} & 71.26\stdv{0.8} & 29.40\stdv{0.7} & 50.31\stdv{0.1} \\
56 & 7 & 2 & 74.97\stdv{0.1} & 77.32\stdv{0.2} & 71.53\stdv{1.1} & 29.60\stdv{0.4} & 50.54\stdv{0.8} \\
56 & 7 & 3 & 75.24\stdv{0.4} & 76.59\stdv{0.2} & 72.44\stdv{1.7} & 29.74\stdv{0.2} & 51.17\stdv{0.9} \\
\bottomrule
\end{tabular}
\end{adjustbox}
\vskip -0.05in
\end{table*}

\Cref{table:ablation} presents the ablation results for patch size, $K_1$, and $K_2$ in CUB-200-2011 dataset. While the accuracy and precision remain relatively consistent across all configurations, localization metrics show variations, particularly with different patch sizes. This is due to the typical scale of the birds within the dataset. A patch size of 56 is generally sufficient to capture which part the patch corresponds to, thus enabling more relevant concepts included in top-$K_1$ concepts.
In contrast, patch size of 32 is relatively small, making it difficult for the CLIP score to accurately reflect the part associated with the patch, leading to decreased localization performance.

There appears to be no significant sensitivity to the values of $K_1$ and $K_2$, which are hyperparameters determining the number of possibly relevant prototypes for a local region.
This indicates that if $K_1$ and $K_2$ are in reasonable range, the model can compactly capture the necessary information.

We selected the configuration that best aligns with both the analysis of the CLIP scores and human intuition. Detailed hyperparameters for each dataset are in Appendix C.

\subsection{Analysis of Prototype Learning}
\subsubsection{Prototype-to-Patch Alignment}

\begin{figure}[ht]
\centering
\hspace*{-0.8cm}
\includegraphics[width=0.4\linewidth,angle=270]{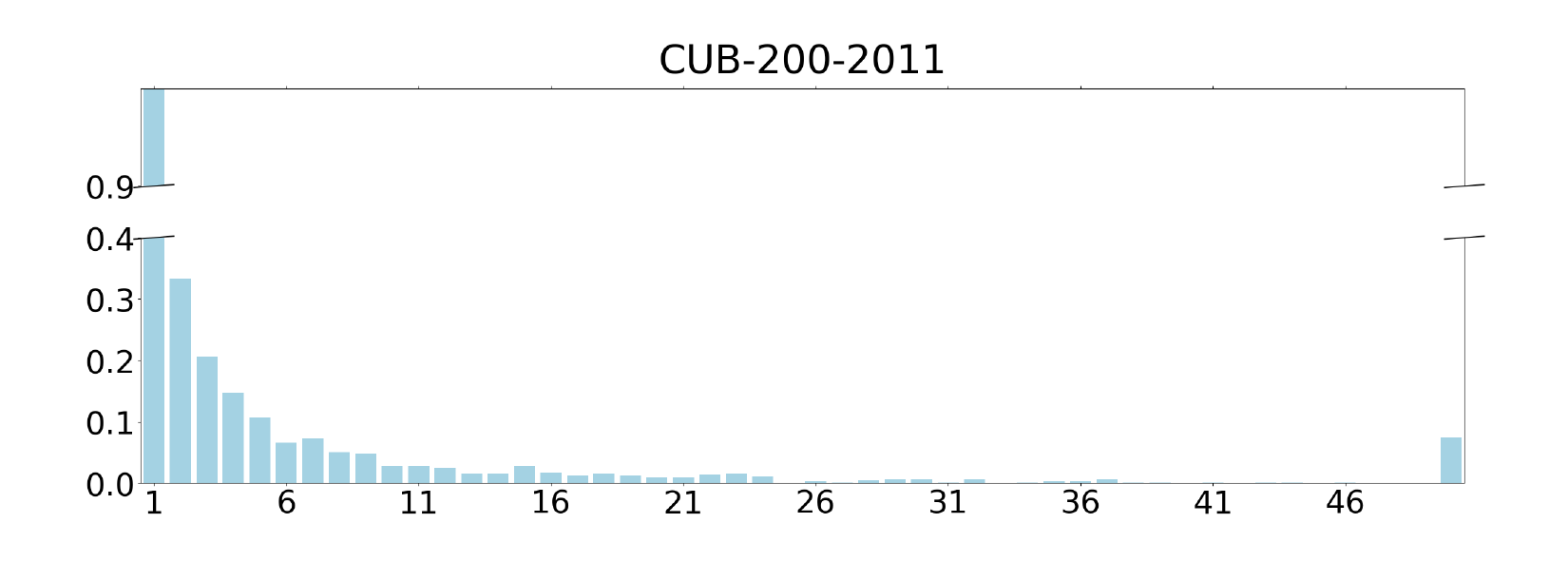}
\caption{Prototype-to-patch alignment results on CUB-200-2011.}
\label{fig:proto-cub}
\end{figure}

\begin{figure}[ht]
\centering
\hspace*{-0.8cm}
\includegraphics[width=0.4\linewidth,angle=270]{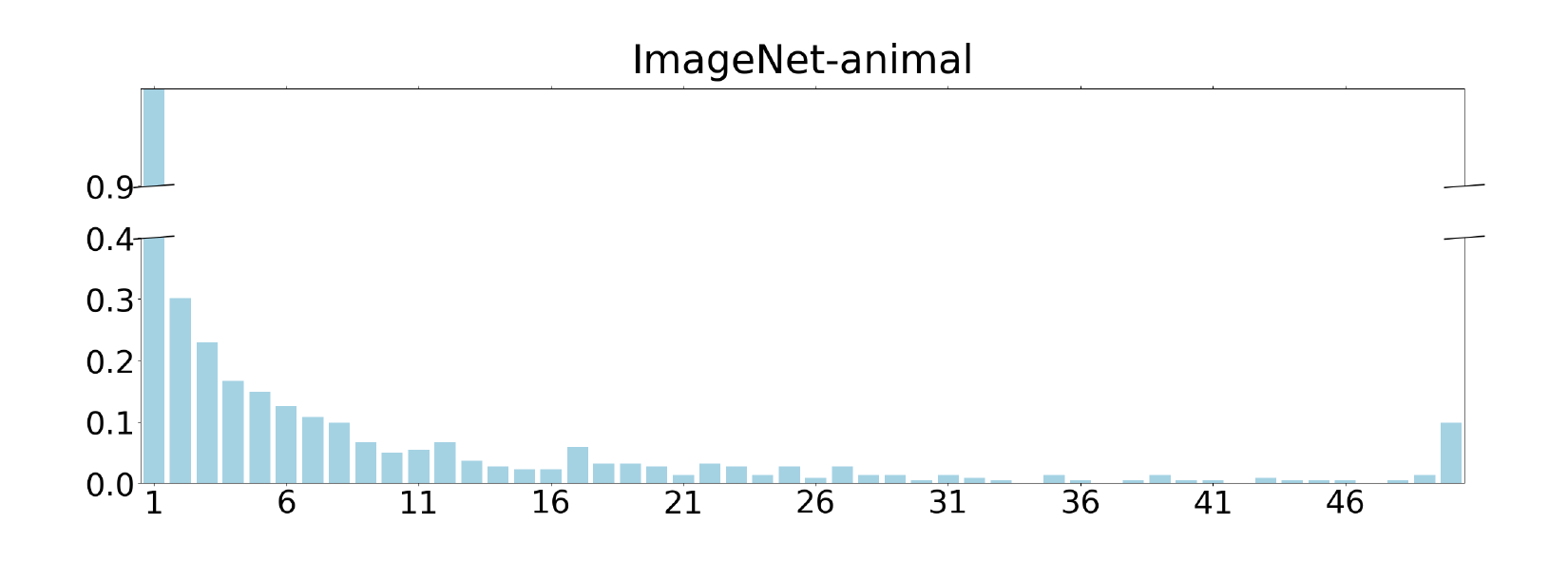}
\caption{Prototype-to-patch alignment results on ImageNet-animal.}
\label{fig:proto-imagenet}
\end{figure}

\begin{figure}[h]
\centering
\hspace*{-0.8cm}
\includegraphics[width=0.4\linewidth,angle=270]{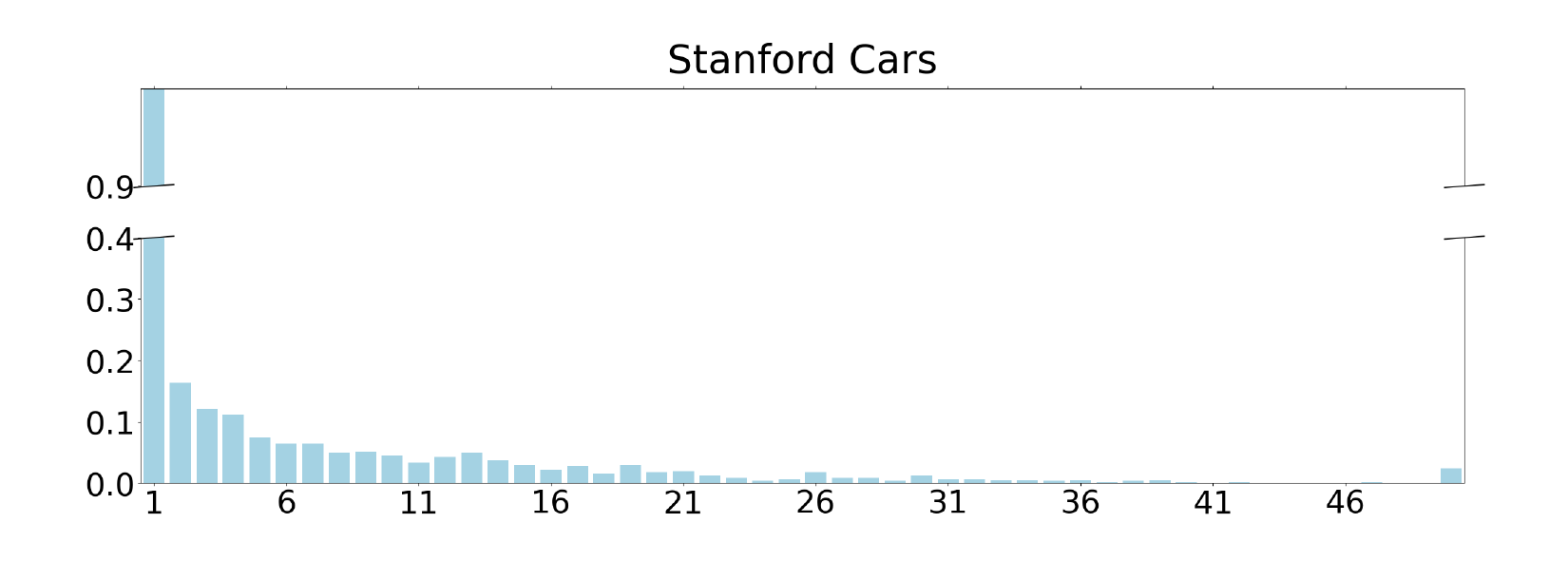}
\caption{Prototype-to-patch alignment results on Stanford Cars.}
\label{fig:proto-cars}
\end{figure}

To validate the prototype learning in our framework, we examined whether the most similar region for each prototype truly aligns with the ground truth region. Specifically, we generated a similarity map ($M^0$ in \Cref{sec:method-arch-c}) between the prototype for an existing concept and the feature map of the input image. Then we sorted all $HW$ similarity values in the descending order, where each location in the feature map is mapped to a value. To match each location in the feature map with input image size, we found the coordinate for each location after bilinear interpolation. Finally, we examined the rank of location in sorted similarity values whose interpolated coordinate is inside the bounding box of corresponding concept. In short, we call it match-rank of the prototype. For example, if the location with max similarity value falls inside the bounding box, the match-rank will be 1 for that prototype. For all annotated concepts, we calculated the match-rank of corresponding prototypes. Fitzpatrick17k dataset is not contained in this evaluation since their annotation strategy is different from other datasets. 

\Cref{fig:proto-cub} to \Cref{fig:proto-cars} demonstrates the results for each dataset. The $x$-axis represents the match-ranks, and the $y$-axis represents the number of prototypes having each match-rank. The last element in $x$-axis denotes the case when none of $HW$ locations falls inside the bounding box. The results show that most prototypes have match-rank of one, with the majority of the remaining prototypes exhibiting low match ranks. This indicates that the prototypes are well-aligned with corresponding regions. 
Since we analyzed only $HW$ locations in the whole image, it could be challenging for them to fall precisely within the bounding box when it is excessively small, resulting in some prototypes have no match-rank.

\begin{table}[h]
\centering
\caption{Patch-to-prototype alignment results.}
\label{table:prototype}
\vskip 0.15in
\begin{adjustbox}{width=0.9\linewidth}%
\begin{tabular}{ccc}
\toprule
CUB-200-2011 & ImageNet-animal & Stanford Cars \\ \midrule 
66.48\stdv{2.0} & 70.24\stdv{0.7} & 83.97\stdv{4.2} \\ 
\bottomrule
\end{tabular}
\end{adjustbox}
\vskip -0.1in
\end{table}

\subsubsection{Patch-to-Prototype Alignment}
We also examined the alignment of prototypes with each local region. For this, we used the point annotations described in \Cref{sec:evaluation-protocol}. Specifically, for patches containing each point annotation, we selected ten prototypes with the highest similarity to the image features at that location. 
We then checked whether any of these prototypes corresponds to a concept that contains the annotated part in its text. For example, if a patch contains a point annotated as ``wing'', we examined whether one of the top prototypes corresponds to the ``wing'' part, such as ``solid yellow wing''. Finally, we calculated the ratio of annotated points having a matching prototype to the total number of annotated points in the image. Fitzpatrick17k dataset is not contained in this evaluation since their concepts are not part-wise. 

\Cref{table:prototype} presents the results, demonstrating that a high proportion of patches are aligned with their corresponding prototypes. For CUB-200-2011 and ImageNet-animal, the ratio is lower compared to Stanford Cars. This discrepancy arises from the presence of concepts that describe the patch accurately but do not explicitly mention the annotated part in their text (e.g., ``solid orange feather'' or ``brown silky fur''). Since these concepts are excluded under this protocol, they are not considered as matches. In contrast, Stanford Cars, which only includes concepts related to discriminative parts (e.g., bumper, grille), achieves higher scores.

\begin{figure*}[ht]
\centering
\includegraphics[width=\textwidth]{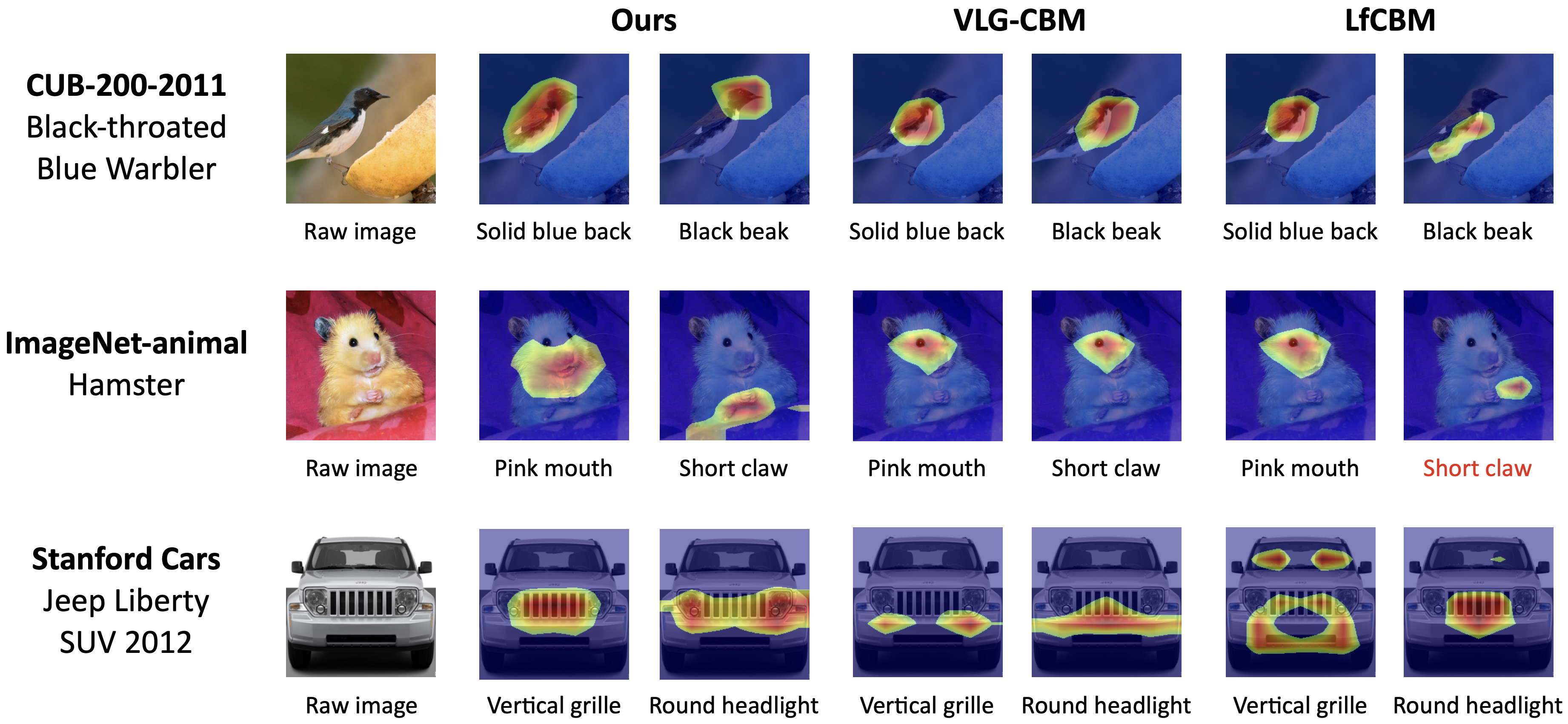}
\caption{Qualitative results of \ours{} and baselines. The leftmost picture shows the input image for each dataset, while the remaining pictures depict the activated regions for the concept indicated below each image. %
Colors closer to red represent stronger activation. A concept highlighted in red indicates negative concept scores, implying that the model predicted the concept to be absent in the image. }
\label{fig:quali}
\end{figure*}

\subsection{Qualitative Analysis}
\Cref{fig:quali} presents the qualitative localization results for three datasets across \ours{} and baselines. The activated area in each image is derived from thresholded GradCAM maps, consistent with the localization evaluation protocol in \Cref{sec:evaluation-protocol}. 

The first row shows results on CUB-200-2011 for the ``Black-throated Blue Warbler'' class. 
It is evident that \ours{} appropriately refers to the corresponding regions when predicting the concepts. For instance, when predicting the concept ``solid blue back'', the model activates around the ``back'' region, while it activates the ``beak'' region when predicting the concept ``black beak''. 
In contrast, VLG-CBM and LfCBM fail to predict each concept from proper regions, especially for concept in relatively small regions such as  ``black beak''. 

The second row shows the results on ImageNet-animal for the class ``Hamster''. Although the activated area may appear coarse, \ours{} predominantly highlights the relevant region. For instance, when \ours{} predicts the concept ``pink mouth'', the highest activation, indicated in red, is indeed centered around the mouth. However, VLG-CBM and LfCBM attend to irrelevant regions. For instance, they both activate regions around the eye when predicting the concept ``pink mouth.''

The third row displays the results on Stanford Cars for the `Jeep Liberty SUV 2012' car type. It is apparent that \ours{} consistently utilizes valid concepts that are well-localized in the image. In contrast, concept predictions of VLG-CBM and LfCBM are not properly localized in corresponding regions. 
Additional qualitative examples are provided in Appendix D.6. Extended qualitative analyses are provided in Appendix D.7 and D.8.

%% file: sections/05-limitation.tex
Although our method incorporates prototype learning, it still inherits inherent biases from CLIP. Moreover, because CLIP is not optimized for identifying detailed concepts within an image, using it in its default form may not be the best choice for our framework. Consequently, minimizing the dependence on CLIP represents a promising direction for future work.

%% file: sections/06-conclusion.tex
In this paper, we tackled the issue of the localization of concepts in existing label-free CBMs by introducing a novel framework, \ours{}. Our approach adopts prototype learning and incorporates two additional losses, one guiding the learning of prototypes and the other enhancing the alignment between the concepts and corresponding local regions.
Experimental results demonstrate that \ours{} significantly improves the localization ability of concept predictions while maintaining a strong explanation capability.
We validated our prototype learning approach by analyzing the relationship between prototypes and local regions.

%% file: supp.tex
\title{Appendix for Locality-aware Concept Bottleneck Model}

\subsection{Concept Set Generation}
\label{appendix:concept-set-generation}

For datasets except Fitzpatrick17k, we generated compositional concepts that can be shared across classes in the form of $<$attribute part$>$. This approach offers two advantages. First, as the concepts are shared across classes, it enables better utilization of representations of local regions that are visually associated with the same concept. Second, compositional concepts are more suitable for the concepts to be localized, allowing more precise alignment between concepts and local regions.
The detailed procedure and prompts for GPT-4o\citep{gpt4} are outlined below. All prompts were refined based on the quality of the GPT-4o responses.
\subsubsection{CUB-200-2011}
For CUB-200-2011 \citep{cub}, we first manually identified parts useful for distinguishing between bird species, including: [\texttt{eye, beak, head, neck, chest, wing, belly, back, leg, feet, tail, feather}]. Next, we prompted GPT-4 to generate possible colors, shapes, and patterns for each part. The detailed prompts are provided below:

\vspace{5pt}
\noindent\fbox{%
    \parbox{\textwidth}{%
        \texttt{What can be the color of a bird's \{part\}?}
        
        \texttt{What are possible patterns of a bird's \{part\}?}
        
        \texttt{What are possible adjectives describing a visual shape of a bird's \{part\}?}
}
}
\vspace{5pt}

For each prompt, we selected a subset of parts appropriate for that specific prompt. For instance, since \texttt{eye} cannot have a pattern, it was excluded from the second prompt.

Next, we combined the attributes from the prompt responses with the identified body parts. Color and shape were treated as separate attributes, while pattern and color were combined if a body part could exhibit both. For example, the concept set for \texttt{head} includes both \texttt{striped black head} and \texttt{rounded head}. 

Finally, we prompted GPT-4o to align the possible concepts for each part with a specific bird class.

\vspace{5pt}
\noindent\fbox{%
    \parbox{\textwidth}{%
    \texttt{Find a correct description for the \{part\} of \{class\} from the following list: \{concepts\}}
}
}
\vspace{5pt}

For example, identifying concepts for the class ``Black Footed Albatross'' for the part 'tail' is as follows:

\vspace{5pt}
\noindent\fbox{%
    \parbox{\textwidth}{%
    \texttt{Find a correct description for the tail of Black Footed Albatross from the following list: [forked tail, rounded tail}, $\cdots$, \texttt{barred white tail}]}
}
\vspace{5pt}

The final concept set $C$ is formed by taking the union of the concepts for all classes.
Concepts that are not aligned with any class were excluded from the final concept set $C$.

\subsubsection{ImageNet-animal}
For ImageNet-animal \citep{imagenet}, we employed a method similar to that used for CUB-200-2011 for concept set generation. However, ImageNet-animal required a more fine-grained procedure due to its diverse range of animal categories.

First, we prompted GPT-4o to classify each animal class into six categories and asked for the possible parts that could exist within each category.

\vspace{5pt}
\noindent\fbox{%
    \parbox{\textwidth}{%
    \texttt{Where does the \{class\} belong to?: [mammal, aquatic vertebrate, bird, reptile, amphibian, invertebrate].}

    \texttt{List all visual parts that can be used to distinguish the species between \{category\}}
}
}
\vspace{5pt}

Then, for each category, we queried GPT-4o to list the possible concepts associated with each part.

\vspace{5pt}
\noindent\fbox{%
    \parbox{\textwidth}{%
    \texttt{What features (e.g.: pattern, shape) can the \{category\}'s \{part\} have?} 
    
    \texttt{Refer to following species but do not be restricted to them : [\{classes in the category\}].} 
    
    \texttt{Answer should be in the form of '(feature) (\{part\})'.}}
}
\vspace{5pt}

For example, the query for generating concepts related to the limb of a mammal is as follows:

\vspace{5pt}
\noindent\fbox{%
    \parbox{\textwidth}{%
    \texttt{What features (e.g.: pattern, shape) can the mammal's limb have?}
    
    \texttt{Refer to following species but do not be restricted to them : [English Springer, Welsh Springer Spaniel,} $\cdots$, \texttt{Tusker}].
    
    \texttt{Answer should be in the form of '(feature) (limb)'.}
}
}
\vspace{5pt}

Finally, we extracted concepts for each class by prompting GPT-4o with the following query:

\vspace{5pt}
\noindent\fbox{%
    \parbox{\textwidth}{%
    \texttt{Concepts: \{concepts for specific part of each category\}}
    
    \texttt{From Concepts, choose the closest description of \{part\} in \{class\} if \{part\} typically exists in \{class\}.}
    
    \texttt{Choose the closest color for the \{part\} of \{class\} from the following: \{colors\}}
    
    \texttt{Concatenate the description and the color, and return in the form:}
    
    \texttt{(color) (description)}
    
    \texttt{If there is no consistent color within the species, return 'none' for color.}
    
    \texttt{Ask yourself again if the concatenated expression is right.} 
    
    \texttt{If not, choose again.}
    
    \texttt{Examples}
    
    \texttt{1. Jellyfish}
    
   \texttt{ -> Since jellyfishes have different colors, you should return 'none smooth tantacle' when the description is smooth tantacle.}
    
    \texttt{2. Wombat}
    
    \texttt{-> Suppose you initially chose 'pink rounded nose' for wombat.} 
    
    \texttt{Ask yourself again, 'Is the nose of a wombat pink and rounded?'}
    
    \texttt{If the answer is yes, return the concatenated description.}
    
    \texttt{If the answer is no, choose another description or color.}
    
    \texttt{In this example, since a wombat's nose is typically brown, you should return 'brown rounded nose'.}
}
}
\vspace{5pt}

The final concept set $C$ is formed by taking the union of the concepts for all classes.
Concepts that are not aligned with any class were excluded from the final concept set $C$.

\subsubsection{Stanford Cars}

For Stanford Cars \citep{cars}, we collected exterior visual concepts that are useful for distinguishing different car models by querying GPT-4o with the following prompt. \{\texttt{classes}\} is the whole classes in the dataset.

\vspace{5pt}
\noindent\fbox{%
    \parbox{\textwidth}{%
    \texttt{Provide a bulleted list of exterior visual features of car models that can be used to distinguish their images.}
    
    \texttt{The features should cover as many car types as possible.}
    
    \texttt{The features should be describable within the context of the image.}

    \texttt{For example, <halogen headlight> is not appropriate.}
    
    \texttt{Each feature should follow the format <attribute> <part>,}
    
    \texttt{and the <attribute> should be diverse enough to describe the <part> of a wide range of car images.}
    
    \texttt{<attribute> can contain shapes or colors.}

    \texttt{Example: <orange> <headlight>}
    
    \texttt{Give at least 10 attributes for each part.}

    \texttt{Refer to the following cars, but do not be restricted to them: \{classes\}}
    }
}
\vspace{5pt}

Then we assigned the concepts to each class by prompting GPT-4o the following:

\vspace{5pt}
\noindent\fbox{%
    \parbox{\textwidth}{%
    \texttt{This is the list of visual concepts that can appear in cars. [\{concepts\}]}
    
    \texttt{What concepts are included in the car model: \{class\}?}
    
    \texttt{Provide bulleted list of the included concepts.}
    }
}
\vspace{5pt}

The final concept set $C$ is formed by taking the union of the concepts for all classes.
Concepts that are not aligned with any class were excluded from the final concept set $C$.

\subsubsection{Fitzpatrick17k}
For Fitzpatrick17k \citep{fitz}, we maintained the form of $<$attribute part$>$ for the concepts, while generating them class-wise. We adopted a different concept generation strategy for Fitzpatrick17k, as there are no explicit common parts associated with skin disease lesions. Nonetheless, the generated concepts can still be shared across classes due to their generality. Detailed prompts for concept set generation are descirbed below:

\vspace{5pt}
\noindent\fbox{%
    \parbox{\textwidth}{%
    \texttt{Give the list of visual features appearing in \{class\} images.}
    
    \texttt{The features should not be semantically duplicated.}
    
    \texttt{Easy words are preferred than professional words.}
    
    \texttt{For example, 'papule' can be replaced by 'bump'.}
    
    \texttt{Features should have the format of <adjective> <noun>.}
    
    \texttt{Example: reddish skin}
    }
}
\vspace{5pt}

\subsection{Experimental Details}
\label{appendix:experimental-details}

\subsubsection{Precision Evaluation Details}
\label{appendix:experimental-details-precision}
For evaluating precision, we sampled 1,000 images for each dataset. Then we prompted GPT-4o to determine whether the top-k concepts predicted by the models truly exist in the given image. When obtaining top-$k$ concepts, we observed 20 concepts with highest contributions, since we determined that concepts whose rank is outside 20 have less meanings. The value of $k$ is 10 for all datasets except ImageNet-animal, where $k$ is 5. We used smaller $k$ for ImageNet-animal since the number of activate concepts, i.e., concepts with positive contributions and positive concept scores, was frequently smaller than 5 in some baselines.

We only considered concepts where the concept contributions are positive. For baselines that allow negative weights between concepts and class predictions, concepts with negative scores could appear in the top-$k$. Therefore, we excluded these concepts with negative scores before selecting the top-$k$. 

Different prompts were used between datasets to accommodate their specific characteristics. The common template for the prompts is as follows, where \texttt{\{category\}} represents the category of objects in the dataset, and \texttt{\{concept list\}} refers to the top-$k$ concepts predicted by the model:

\vspace{5pt}
\noindent\fbox{%
    \parbox{\textwidth}{%
    \texttt{Answer to my question asking whether each concept exists in the image.}
    
    \texttt{IMPORTANT: avoid being strict.}
    
    \texttt{Even if the match is not exact, consider partial matches valid.}
    
    \texttt{Always prioritize flexible, inclusive judgments over strict ones unless explicitly told otherwise. If unsure, err on the side of considering the concept present.}
    
    \texttt{Question: Does \{concept list\} appear in any part of this \{category\}'s image, when keeping the above instruction in mind?}
    
    \texttt{Provide an explanation for your decision and also a verification whether you followed the IMPORTANT instruction. If not, revise your answer accordingly.}
    
    \texttt{Finally give the yes or no answer, Separating it by '/' in the '<>'.}
    
    \texttt{Example: <yes/no/no/yes/no>}
}
}
\vspace{5pt}

We used this prompt to instruct GPT-4o to make inclusive decisions, as this approach reduces the randomness of its responses. 

The whole prompt for CUB-200-2011 and ImageNet-animal is as follows, while the \texttt{\{category\}} is \texttt{bird} for CUB-200-2011 and \texttt{animal} for ImaegeNet-animal:

\vspace{5pt}
\noindent\fbox{%
    \parbox{\textwidth}{%
    \texttt{Answer to my question asking whether each concept exists in the image.}
    
    \texttt{IMPORTANT: avoid being strict.}
    
    \texttt{Even if the match is not exact, consider partial matches valid.}

    \texttt{Example 1: If a tail is mostly white with slight gray but does not perfectly match the concept 'solid white,' it should still be considered present if it reasonably aligns with the concept.}

    \texttt{Example 2: If there is a coloring of grayish brown in the image, both 'gray feather' and 'brown feather' should be considered present, too.}
    
    \texttt{Always prioritize flexible, inclusive judgments over strict ones unless explicitly told otherwise.}
    
    \texttt{If unsure, err on the side of considering the concept present.}
    
    \texttt{Question: Does \{concept list\} appear in any part of this \{category\}'s image, when keeping the above instruction in mind?}
    
    \texttt{Provide an explanation for your decision and also a verification whether you followed the IMPORTANT instruction. If not, revise your answer accordingly.}
    
    \texttt{Finally give the yes or no answer, Separating it by '/' in the '<>'.}
    
    \texttt{Example: <yes/no/no/yes/no>}
}
}
\vspace{5pt}

For Fitzpatrick17k, we added a caution to the prompt. The whole prompt is provided below:

\vspace{5pt}
\noindent\fbox{%
    \parbox{\textwidth}{%
    \texttt{Answer to my question asking whether each concept exists in the image.}
    
    \texttt{IMPORTANT: avoid being strict.}
    
    \texttt{Even if the match is not exact, consider partial matches valid.}
    
    \texttt{Always prioritize flexible, inclusive judgments over strict ones unless explicitly told otherwise.}
    
    \texttt{Keep in mind that the images are of skin diseases-you should interpret all concepts in terms of skin-related features.}
    
    \texttt{If unsure, err on the side of considering the concept present.}
    
    \texttt{Question: Does \{concept list\} appear in any part of this lesion's image, when keeping the above instruction in mind?}
    
    \texttt{Provide an explanation for your decision and also a verification whether you followed the IMPORTANT instruction.}
    
    \texttt{If not, revise your answer accordingly.}
    
    \texttt{Finally give the yes or no answer, Separating it by '/' in the '<>'.}
    
    \texttt{Example: <yes/no/no/yes/no>}
}
}
\vspace{5pt}

For Stanford Cars, we used the default template. 

\subsubsection{Localization Evaluation Details}
\label{appendix:experimental-details-localization}
We generated GradCAM map using publicly available code.\footnote{\url{https://github.com/jacobgil/pytorch-grad-cam.git}} Then we applied threshold of 0.5 after dividing the GradCAM map with its max value. The number of images we used for localization evaluation is 300 for each dataset, except Fitzpatrick17k whose number of annotated images is 100.

\subsubsection{Model and Hyperparameter Selection}
We employed early stopping with a patience of 3 for ImageNet-Animal and 5 for the other datasets. The check interval was set to 5 for the baselines and 1 for our method, as the baselines train a single linear layer, whereas our approach entails end-to-end training of the entire framework. We then selected the model with the highest validation accuracy before early stop.

For hyperparameter selection, we performed a grid search of subset of hyperparameters for the baselines and selected the model with the highest validation accuracy. For our method, hyperparameters were primarily selected based on validation accuracy, while additionally considering the effectiveness and reliability of explanations.

\subsection{Implementation Details}
\label{sec:implementation-details}

\subsubsection{Baselines}
\label{sec:implementation-details-baselines}
Here, we provide the details of the baselines, LfCBM, LaBo, and VLG-CBM.
\begin{itemize}[leftmargin=*]
    \item LfCBM \citep{labelfree}: LfCBM sequentially trains the concept prediction layer and classification layer. For the concept prediction layer, they used the cubed cosine similarity between the CLIP score and the predicted concept scores to guide its learning.
    We use the publicly available code provided by the authors.\footnote{\url{https://github.com/Trustworthy-ML-Lab/Label-free-CBM}}
    \item LaBo \citep{labo}: LaBo treats the concept prediction layer as fixed, directly using the calculated CLIP scores as input to the classification layer. 
    We use the publicly available code provided by the authors.\footnote{\url{https://github.com/YueYANG1996/LaBo}}
    \item VLG-CBM \citep{vlgcbm}: VLG-CBM tackles the issue of improper alignment of concepts to the images by employing an open-vocabulary object detection module to identify and annotate the concepts present in the image. These annotations are then used to generate one-hot vectors representing the existence of each concept, which serve as guidance for learning of the concept prediction layer. It sequentially trains concept prediction layer and classification layer, similar to LfCBM.
    Note that although VLG-CBM recently proposed `grounding' concepts in images via an open-domain object detection module, their definition of `grounding'
    pertains to concept annotation, rather than within the concept prediction.  
    We use the publicly available code provided by the authors.\footnote{\url{https://github.com/Trustworthy-ML-Lab/VLG-CBM}}
\end{itemize}

For LaBo, we excluded the concept selection module since our concept sets are not initially created class-wise.
For backbones of the baselines, we used ResNet50 \citep{resnet} pretrained on ImageNet and fine-tuned on each dataset, except for ImageNet-animal, which is already pretrained on the ImageNet dataset. 
We used the same concept set generated at section 3.1 for all baselines. 

\subsubsection{Ours}
For backbones of ours, we used ResNet50 pretrained on the ImageNet and fine-tuned on each dataset for Stanford Cars and Fitzpatrick17k. We used ResNet50 only pretrained on the ImageNet for CUB-200-2011 and ImageNet-animal.

For CLIP, we used CLIP-ViT/16 for both of ours and all baselines.

Hyperparameters we used are in \Cref{table:hyperparameter}. Since patch sizes are larger than 32, which is the patch size when dividing $224\times 224$ image by $7\times 7$ grid, we overlapped the patches. For optimizer, we used AdamW\citep{adamw} for all datasets.

\begin{table*}[t]
\centering
\caption{Hyperparameters}
\label{table:hyperparameter}
\vskip 0.15in
\begin{adjustbox}{width=0.7\textwidth}%
\begin{tabular}{cccccccc}
\toprule
Dataset & $\alpha$ & $\beta$ & Patch size & $K_1$ & $K_2$ & Batch size & Learning rate \\ \midrule
CUB-200-211 & 0.5 & 0.1 & 56 & 7 & 3 & 8 & $5e-5$\\ 
ImageNet-animal & 1.0 & 0.1 & 74 & 5 & 3 & 8 & $1e-5$\\
Stanford Cars & 0.5 & 0.5 & 56 & 7 & 3 & 8 & $5e-5$\\
Fitzpatrick17k & 1.0 & 0.5 & 56 & 15 & 7 & 8 & $5e-5$\\
\bottomrule
\end{tabular}
\end{adjustbox}
\vskip -0.1in
\end{table*}

\begin{table*}[h]
\centering
\caption{Black-box Model Accuracy}
\label{table:blackbox}
\vskip 0.15in
\begin{adjustbox}{width=0.4\textwidth}%
\begin{tabular}{ccc}
\toprule
CUB-200-2011 & Stanford Cars & Fitzpatrick17k\\ \midrule
78.25 & 83.48 & 52.72 \\
\bottomrule
\end{tabular}
\end{adjustbox}
\vskip -0.1in
\end{table*}

\subsection{Additional Experiments}
\label{appendix:additional-experiments}
\subsubsection{Black-box Model Accuracy}
\label{appendix:additional-experiments-blackbox}
\Cref{table:blackbox} reports the accuracy of the ResNet50 model pretrained on the ImageNet and fine-tuned on each dataset. For input images, we resized them so that the shorter dimension be 224 pixels, followed by a center crop to $224\times 224$, same as the input image size for ours and all baselines. No data augmentation techniques were applied.

For CUB-200-2011, following the setup in \citep{resnet_finetune}, the model was optimized for 90 epochs using SGD with a momentum of 0.9 and a learning rate of 0.001. The learning rate was reduced by a factor of 0.1 every 30 epochs. For Stanford Cars and Fitzpatrick17k, the model was optimized for 100 epochs using AdamW with a learning rate of 0.001, while the learning rate was reduced by 0.1 every 30 epochs. AdamW was chosen as it produced better validation accuracy.

\begin{table*}[ht]
\centering
\caption{Results on CUB-200-2011}
\label{table:concept-score-cub}
\vskip 0.15in
\begin{adjustbox}{width=0.6\textwidth}%
\begin{tabular}{lcccc}
\toprule
& \multicolumn{3}{c}{Localization} &  Concept Evaluation \\ \cmidrule(lr){2-4} \cmidrule(lr){5-5}
Model & Inclusion & mIoU & REP & Precision \\ \midrule
LfCBM & 66.45\stdv{1.0} & 29.64\stdv{0.3} & 39.16\stdv{0.2} & 46.74\stdv{0.6} \\
LaBo & - & - & - & 45.96\stdv{0.3}\\
VLG-CBM & 52.55\stdv{1.5} & 22.99\stdv{0.3} & 30.59\stdv{0.2} & 64.26\stdv{0.4}\\
\ours{} (Ours) & \textbf{76.46}\stdv{2.2} & \textbf{31.22}\stdv{0.5} & \textbf{53.72}\stdv{0.7} & \textbf{67.50}\stdv{0.1}\\
\bottomrule
\end{tabular}
\end{adjustbox}
\vskip -0.1in
\end{table*}

\begin{table*}[ht]
\centering
\caption{Results on ImageNet-animal}
\label{table:concept-score-imagenet}
\vskip 0.15in
\begin{adjustbox}{width=0.6\textwidth}%
\begin{tabular}{lcccc}
\toprule
& \multicolumn{3}{c}{Localization} & Concept Evaluation\\ \cmidrule(lr){2-4} \cmidrule(lr){5-5}
Model & Inclusion & mIoU & REP & Precision\\ \midrule
LfCBM & 58.44\stdv{1.6} & 26.45\stdv{0.5} & 40.97\stdv{0.3} & 63.35\stdv{0.7} \\
LaBo & - & - & - & 64.49\stdv{0.4} \\
VLG-CBM & 54.53\stdv{1.0} & 24.56\stdv{0.2} & 40.82\stdv{0.3} & 73.16\stdv{0.5}  \\
\ours{} (Ours) & \textbf{82.62}\stdv{1.2} & \textbf{28.44}\stdv{0.4} & \textbf{70.15}\stdv{0.6} & \textbf{73.68}\stdv{0.6} \\
\bottomrule
\end{tabular}
\end{adjustbox}
\vskip -0.1in
\end{table*}

\begin{table*}[ht]
\centering
\caption{Results on Stanford Cars}
\label{table:concept-score-cars}
\vskip 0.15in
\begin{adjustbox}{width=0.6\textwidth}%
\begin{tabular}{lcccc}
\toprule
& \multicolumn{3}{c}{Localization} &  Concept Evaluation\\ \cmidrule(lr){2-4} \cmidrule(lr){5-5}
Model & Inclusion & mIoU & REP & Precision \\ \midrule
LfCBM & 39.18\stdv{1.6} & 18.63\stdv{0.3} & 27.73\stdv{0.1} & 54.60\stdv{0.8} \\
LaBo & - & - & - & 62.24\stdv{0.2} \\
VLG-CBM & 21.78\stdv{0.6}& 11.68\stdv{0.2}& 15.88\stdv{0.2}& \textbf{65.80}\stdv{0.6} \\
\ours{} (Ours) & \textbf{72.41}\stdv{0.6} & \textbf{30.89}\stdv{0.1} & \textbf{53.26}\stdv{0.4}& 60.62\stdv{1.0} \\
\bottomrule
\end{tabular}
\end{adjustbox}
\vskip -0.1in
\end{table*}

\begin{table*}[ht]
\centering
\caption{Results on Fitzpatrick17k}
\label{table:concept-score-fitz17k}
\vskip 0.15in
\begin{adjustbox}{width=0.6\textwidth}%
\begin{tabular}{lcccc}
\toprule
& \multicolumn{3}{c}{Localization} & Concept Evaluation\\ \cmidrule(lr){2-4} \cmidrule(lr){5-5}
Model & Inclusion & mIoU & REP & Precision \\ \midrule
LfCBM & \textbf{87.52}\stdv{2.8} & 36.81\stdv{2.4} & 48.34\stdv{3.7} & 45.70\stdv{0.6} \\
LaBo & - & - & - & \textbf{53.74}\stdv{0.4} \\
VLG-CBM & 60.94\stdv{2.4} & 21.44\stdv{0.8} & 26.01\stdv{1.2} & 46.20\stdv{0.7} \\
\ours{} (Ours) & \underline{84.41}\stdv{0.3} & \textbf{37.34}\stdv{1.2} & \textbf{51.09}\stdv{1.0} & \underline{51.78}\stdv{2.2} \\
\bottomrule
\end{tabular}
\end{adjustbox}
\vskip -0.1in
\end{table*}

\subsubsection{Experiments on Concept Scores}
\label{appendix:additional-experiments-conceptscores}
\Cref{table:concept-score-cub} to \Cref{table:concept-score-fitz17k} present the precision and localization evaluation results based solely on concept scores, without multiplying them with classification weights. Since LaBo does not train a concept prediction layer and directly uses raw CLIP scores as concept scores, the standard deviation reflects the degree of randomness in our GPT-based evaluation. Thus the small standard deviations indicate that the randomness of our evaluation method is reasonably low. The results of \ours{} and other baselines show that \ours{} is also robust in terms of the concept scores. 

\subsubsection{Recall Evaluation}
\label{appendix:additional-experiments-recall}
\label{appendix:recall}

\begin{table*}[ht]
\centering
\caption{Results on Recall Evaluation}
\label{table:recall}
\vskip 0.15in
\begin{adjustbox}{width=0.7\textwidth}%
\begin{tabular}{lcccc}
\toprule
Model & CUB-200-2011 & ImageNet-animal & StanfordCars & Fitzpatrick17k \\ \midrule 
LfCBM & 14.06\stdv{0.3} & 6.55\stdv{0.1} & 13.64\stdv{0.3} & 3.99\stdv{0.1} \\
LaBo &  16.63\stdv{0.0} & 3.49\stdv{0.0} & 10.00\stdv{0.0} & 3.93\stdv{0.0} \\
VLG-CBM & \underline{39.51}\stdv{0.2} & \textbf{48.66}\stdv{0.1} & \underline{39.85}\stdv{0.3} & \textbf{20.84}\stdv{0.2} \\
\ours{} (Ours) & \textbf{41.46}\stdv{1.1} & \underline{45.30}\stdv{0.4} & \textbf{42.90}\stdv{0.9} & \underline{20.62}\stdv{0.6} \\
\bottomrule
\end{tabular}
\end{adjustbox}
\vskip -0.1in
\end{table*}

To examine whether the models catch diverse concepts, we evaluated the recall of predicted concepts for each model. Since it is not feasible to observe every concepts, we selected ground-truth concept candidates from the concepts associated with the label, as described in \Cref{appendix:concept-set-generation}. Then similar to the precision evaluation, we queried these candidate concepts using GPT-4o to determine whether they exist in the image. Recall was then calculated as the ratio of the number of ground-truth concepts that matched the top-$k$ concepts to the total number of ground-truth concepts. The results on recall evaluation in terms of concept scores are in \Cref{table:recall}. 

\Cref{table:recall} shows that \ours{} achieves high recall scores, demonstrating its ability to capture a wide range of necessary concepts. One point to note is that since we evaluated the recall only on a subset of concepts, it does not fully reflect the ability of models to predict diverse concepts. Thus, the fact that \ours{} demonstrates reasonable recall scores is the central aspect, rather than precise comparison with baselines.

\subsubsection{Evaluation on Human Annotations}
we conducted an additional experiment on the entire CUB-200-2011 dataset using its human-annotated concepts, without using GPT. This dataset includes a predefined concept set, annotations specifying which concepts appear in each image, and point annotations marking the locations of bird parts (e.g., head, wing). \Cref{table:gt} demonstrates that \ours{} also exhibit superior performance in this setting.

\begin{table*}[ht]
\centering
\caption{Results on Human Annotations}
\label{table:gt}
\vskip 0.15in
\begin{adjustbox}{width=0.45\textwidth}%
\begin{tabular}{lcccc}
\toprule
Model & Precision & Inclusion & MIoU & REP \\ \midrule 
LfCBM & 23.71 & 59.82 & 29.73 & 42.12 \\
VLG-CBM & 55.17 & 51.24 & 26.58 & 36.64 \\
\ours{} (Ours) & \textbf{62.75}\ & \textbf{75.76} & \textbf{35.37} & \textbf{53.99} \\
\bottomrule
\end{tabular}
\end{adjustbox}
\vskip -0.1in
\end{table*}

\subsubsection{MNIST Toy Experiment on Prototype Learning}
\label{appendix:additional-experiments-mnist}
\begin{figure}[ht]
\centering
\includegraphics[width=\linewidth]{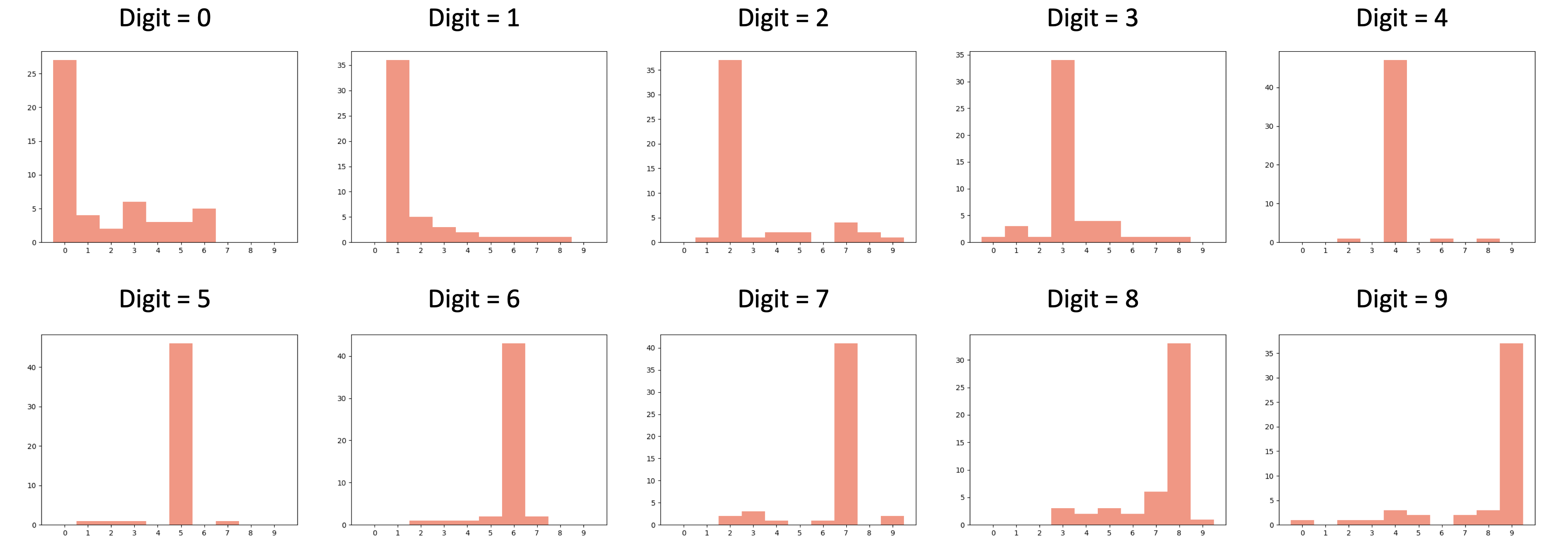}
\caption{Histogram for MNIST experiment.}
\label{fig:mnist-example}
\end{figure}

To validate our prototype learning approach, we conducted a toy experiment on MNIST. First, we created a dataset by concatenating four digit images in a $2\times 2$ grid, resulting in a $56 \times 56$ image. The sorted sequence of the four digits was used as class label for each image, resulting in 10000 classes in total. We generated 50,000 image-label pairs for this dataset.

Next, we set up the experiment to mimic the configuration of our prototype learning. We considered numbers zero to nine as concepts and initialized ten prototypes, each corresponding to one number (i.e., a concept). And we treated each of the four digits in an image as an image patch. To each patch, we assigned two random digits in addition to the ground-truth digit. This setting is as analogous to using top-$K_1$ prototypes selected for each image patch.
We then computed a weighted sum of the prototypes of three concepts: the ground-truth digit and two random digits. 

Similar to our framework, we adopted a classification loss using the weighted sum of the prototypes, and trained the prototypes exclusively with this classification loss. After training, we calculated the similarity between the learned prototypes and 100 random images for each digit from the MNIST test dataset. Finally, we plotted a histogram showing the number of prototypes aligned to each digit image.

\Cref{fig:mnist-example} shows the results for all digits. It demonstrates that most prototypes corresponding to each digit are aligned with the respective digit images, validating the effectiveness of our prototype learninig approach.

\subsection{Additional Qualitative Examples}
\label{appendix:additional-experiments-qualitative}
\begin{figure}[ht]
\centering
\includegraphics[width=0.75\linewidth]{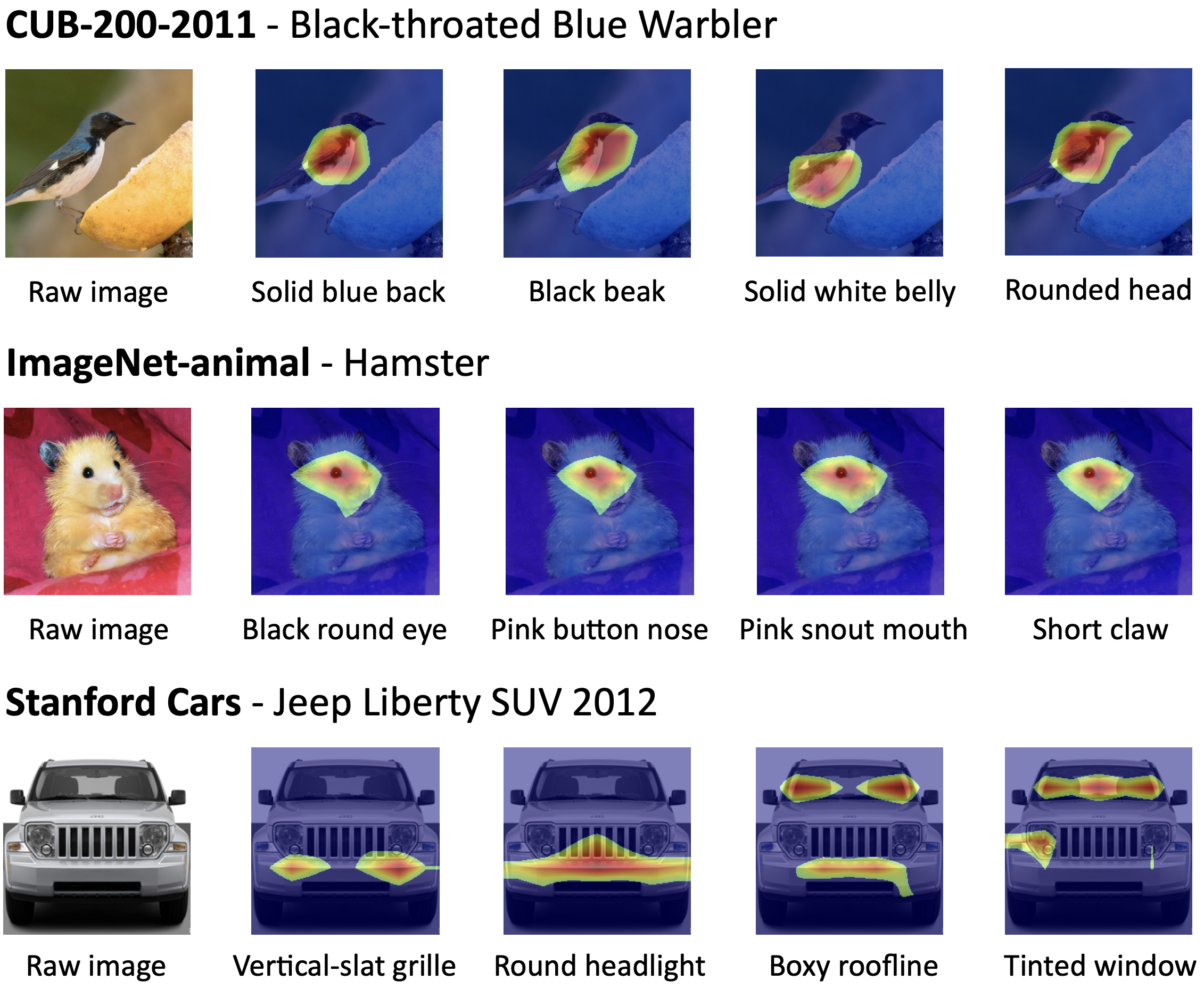}
\caption{Qualitative Results for VLG-CBM.}
\label{fig:quali-vlgcbm}
\end{figure}

\begin{figure}[ht]
\centering
\includegraphics[width=0.75\linewidth]{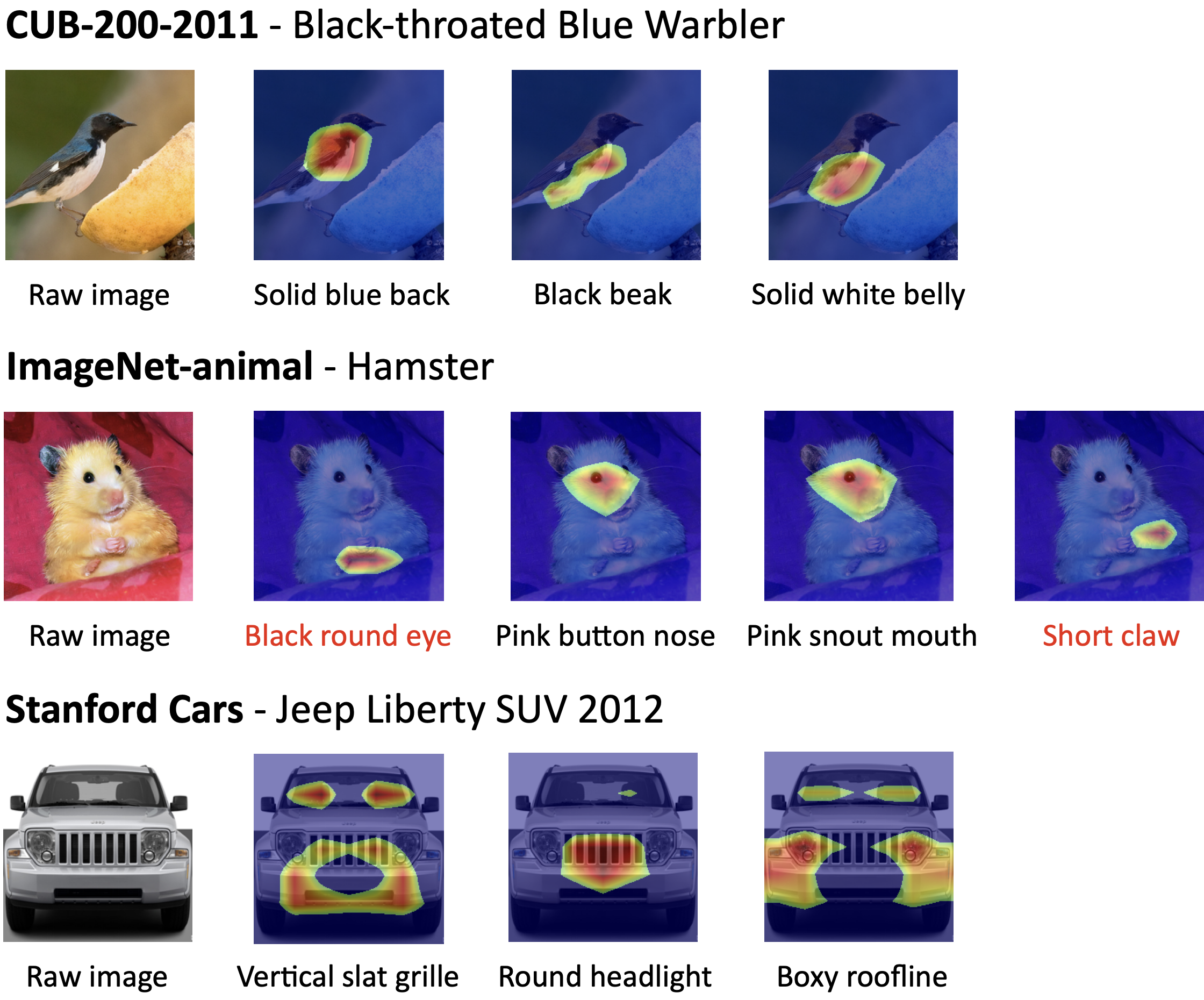}
\caption{Qualitative Results for LfCBM.}
\label{fig:quali-lfcbm}
\end{figure}

\begin{figure}[ht]
\centering
\includegraphics[width=0.75\linewidth]{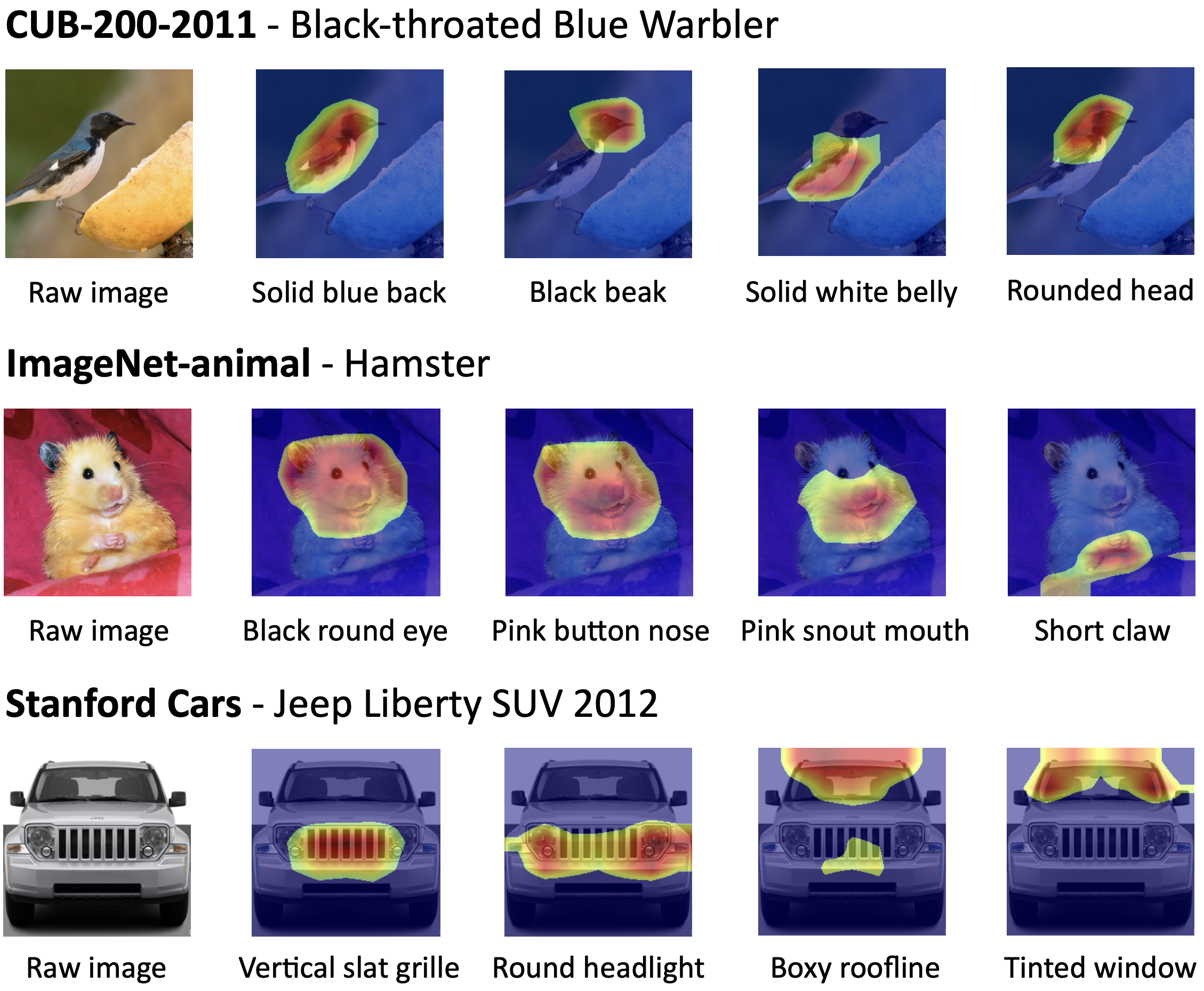}
\caption{Qualitative Results for \ours.}
\label{fig:quali-ours}
\end{figure}

\Cref{fig:quali-vlgcbm} to \Cref{fig:quali-ours} present the results on three datasets, using the same examples as section 4.6. Concepts highlighted in red indicate negative presence scores. LfCBM has fewer examples in CUB-200-2011 and Stanford Cars because the concepts used in other methods were filtered out during its training process.

As shown in each figure, most concept predictions in previous label-free CBMs are not properly localized in the corresponding regions, as their GradCAM maps either activate irrelevant regions or simultaneously activate multiple parts. In contrast, concept prediction of \ours{} are properly localized in corresponding region.

\begin{figure}[ht]
\centering
\includegraphics[width=0.75\linewidth]{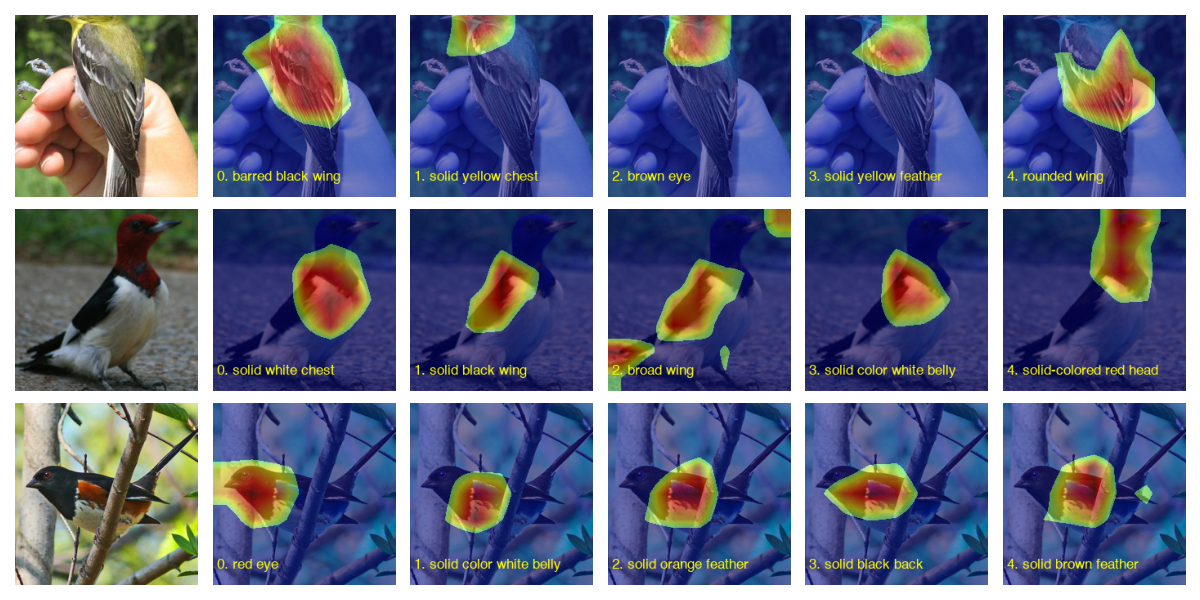}
\caption{Qualitative Results of \ours{} for CUB-200-2011.}
\label{fig:quali-cub}
\end{figure}

\begin{figure}[ht]
\centering
\includegraphics[width=0.75\linewidth]{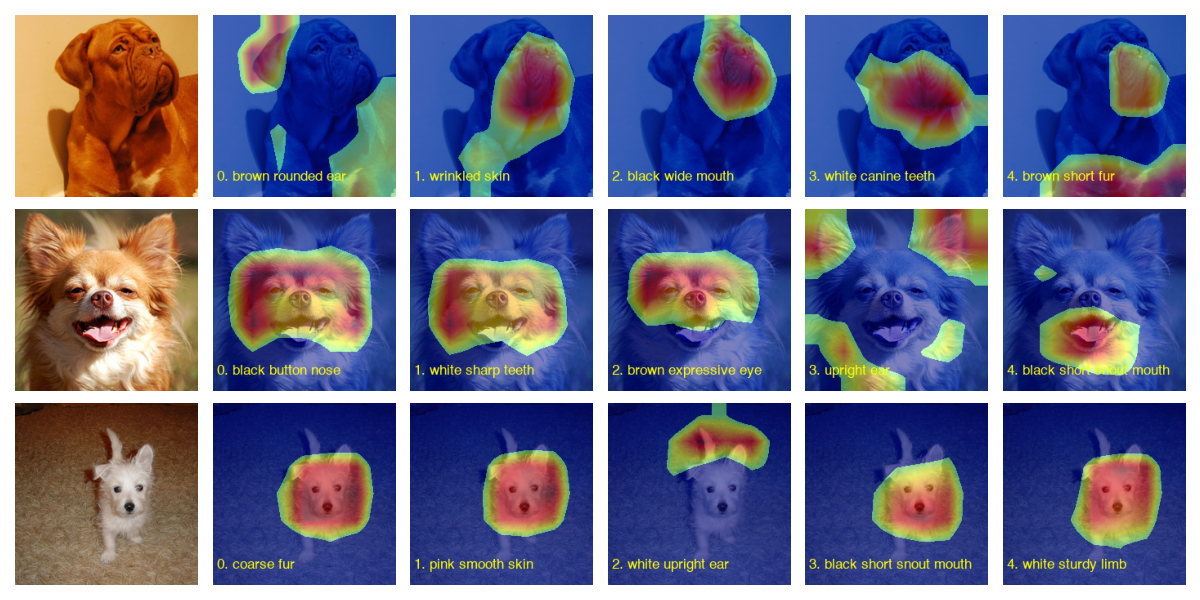}
\caption{Qualitative Results of \ours{} for ImageNet-animal.}
\label{fig:quali-imagenet}
\end{figure}

\begin{figure}[ht]
\centering
\includegraphics[width=0.75\linewidth]{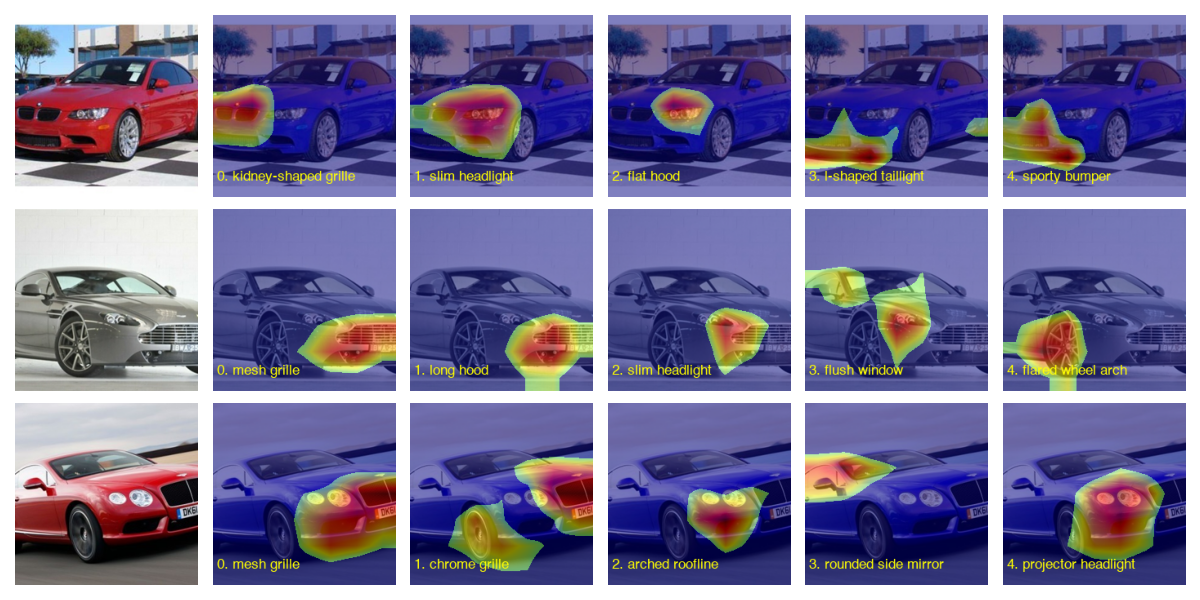}
\caption{Qualitative Results of \ours{} for Stanford Cars.}
\label{fig:quali-cars}
\end{figure}

\Cref{fig:quali-cub}, \Cref{fig:quali-imagenet}, \Cref{fig:quali-cars} are additional qualitative examples, each for CUB-200-2011, ImageNet-animal, and Stanford Cars. Despite minor inaccuracies in some localizations, \ours{} reliably distinguishes the distinct parts of each object.

\subsubsection{Maximum Activating Example Analysis}
\begin{figure}[ht]
\centering
\includegraphics[width=\linewidth]{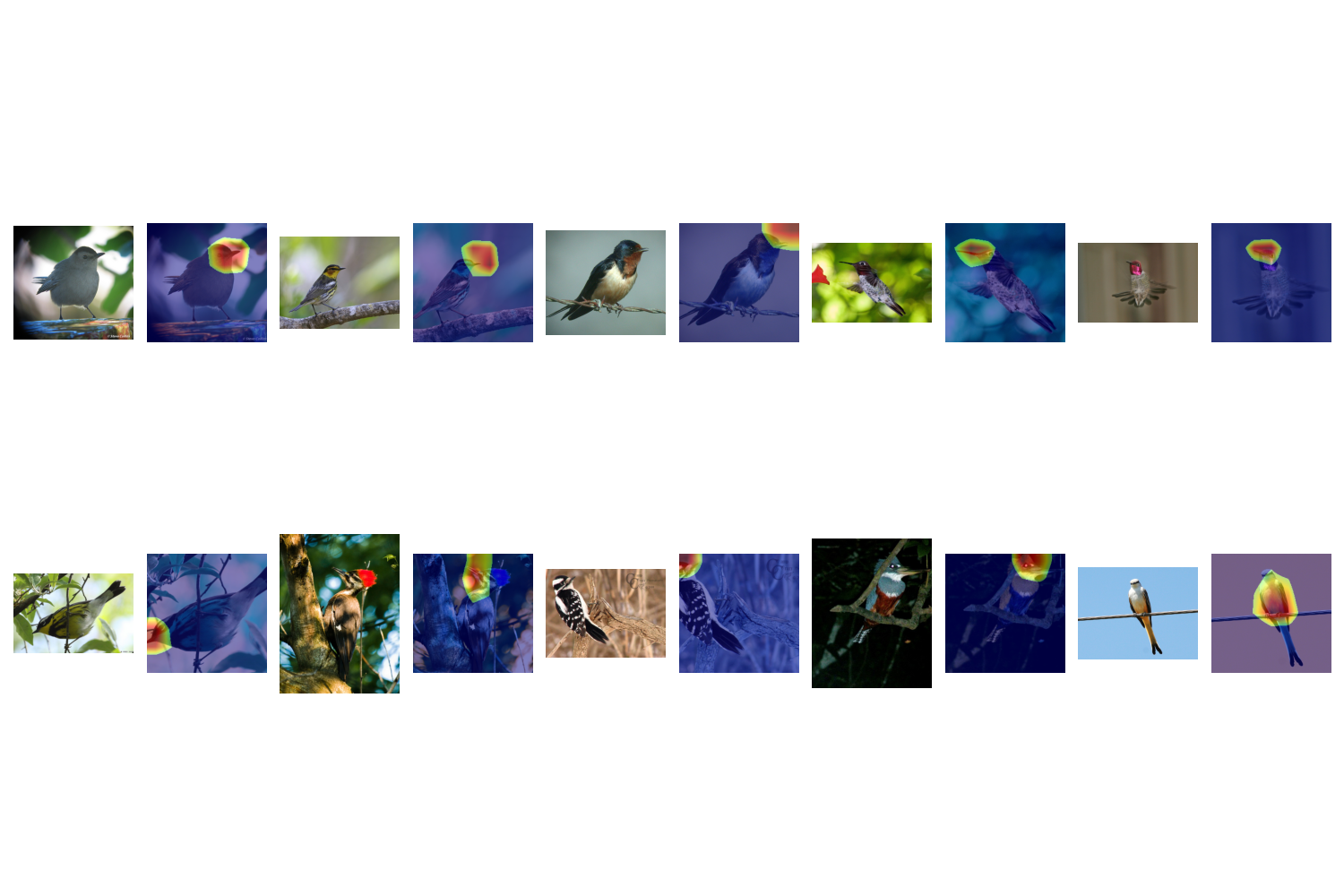}
\caption{Max Activating Examples for Concept Black Bill.}
\label{fig:black-bill}
\end{figure}

\begin{figure}[ht]
\centering
\includegraphics[width=\linewidth]{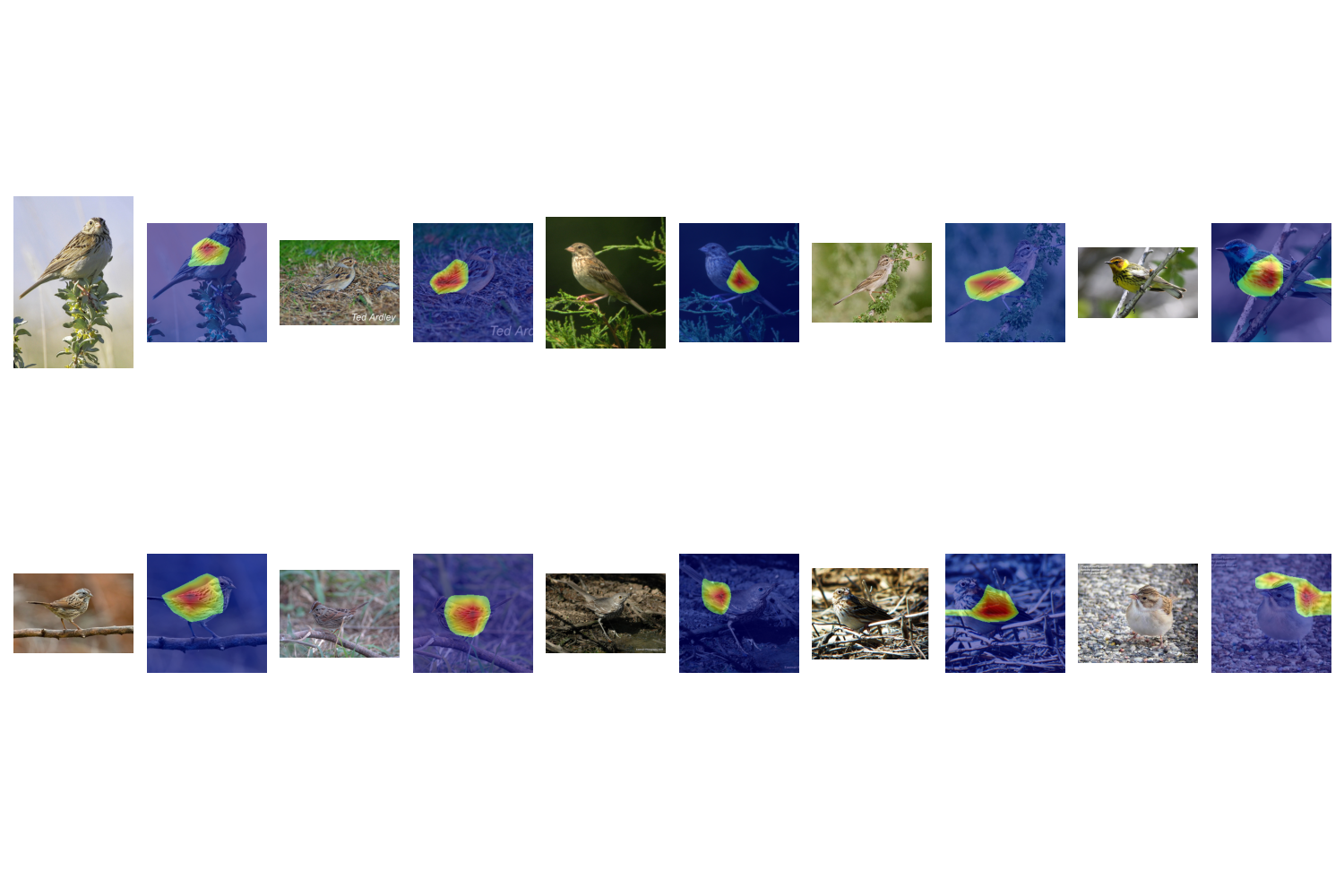}
\caption{Max Activating Examples for Concept Striped Wing.}
\label{fig:striped-wing}
\end{figure}

\begin{figure}[ht]
\centering
\includegraphics[width=\linewidth]{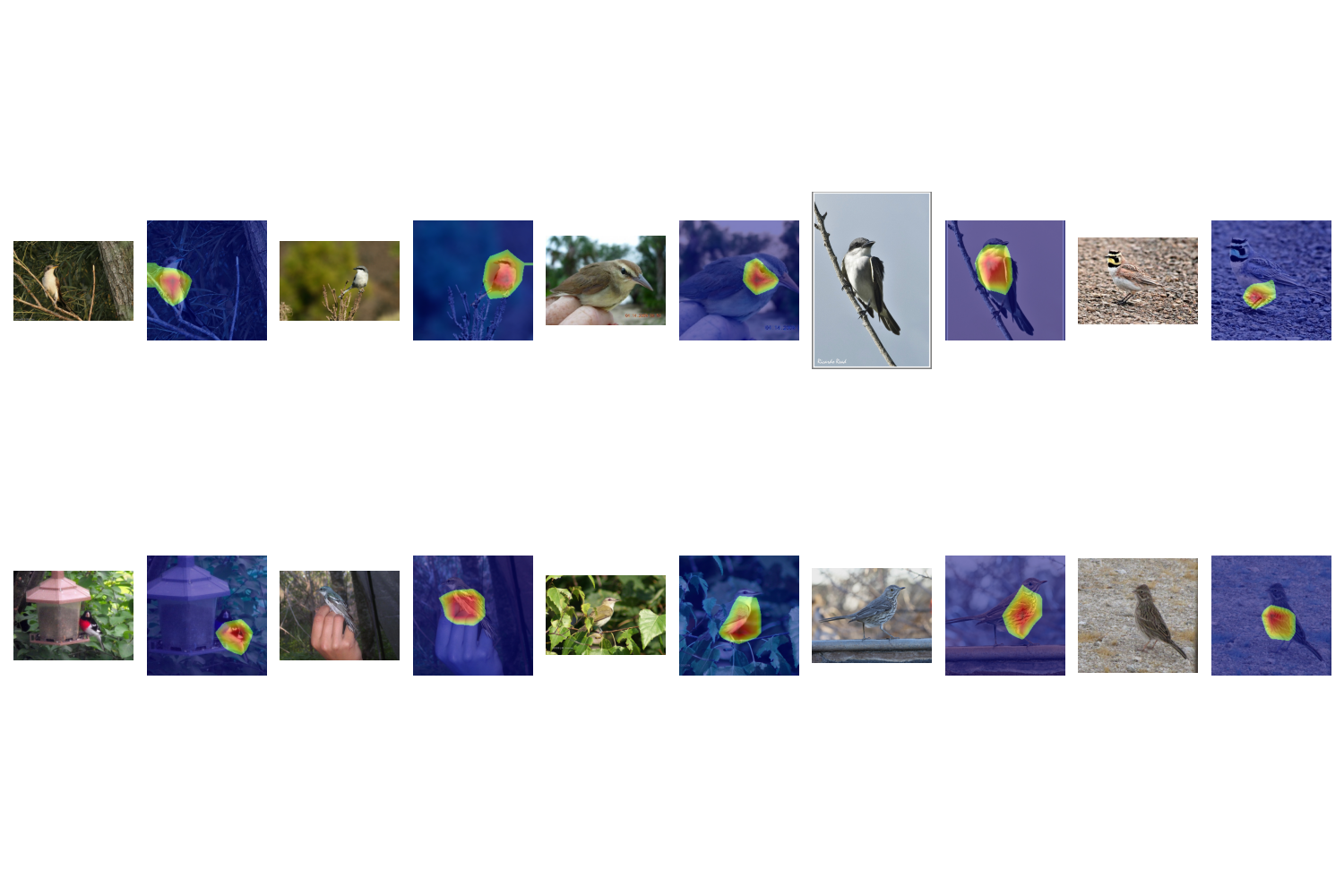}
\caption{Max Activating Examples for Concept White Belly.}
\label{fig:white-belly}
\end{figure}

To observe whether each concept is properly localized globally, we obtained the GradCAM values for each demonstrated concept and randomly selected images where the maximum GradCAM value exceeded 0.999. The results in \Cref{fig:black-bill} to \Cref{fig:white-belly} display the GradCAM maps for each concept. As shown in the results, \ours{} appropriately localizes concepts. 

\subsubsection{Local Explanation Examples}
\begin{figure}[ht]
\centering
\includegraphics[width=\linewidth]{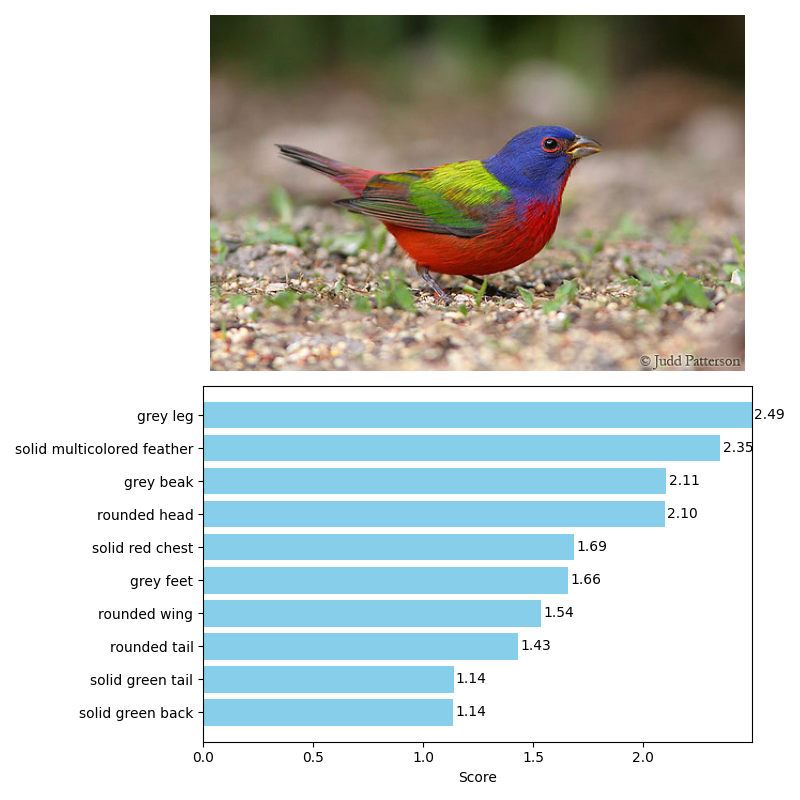}
\caption{Local Explanation Example of CUB-200-2011.}
\label{fig:local_cub_1}
\end{figure}

\begin{figure}[ht]
\centering
\includegraphics[width=\linewidth]{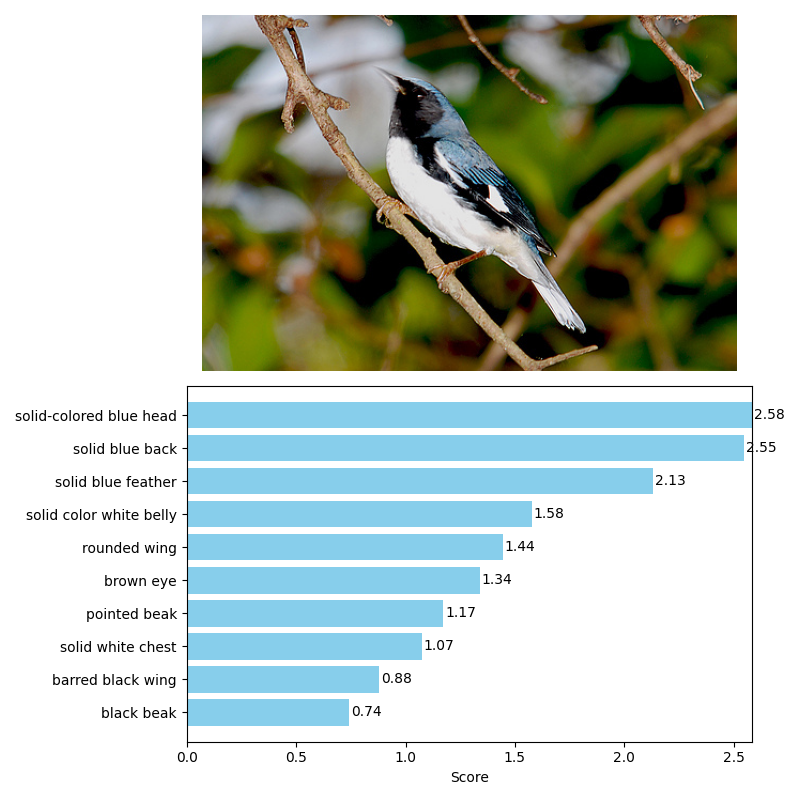}
\caption{Local Explanation Example of CUB-200-2011.}
\label{fig:local_cub_2}
\end{figure}

\begin{figure}[ht]
\centering
\includegraphics[width=\linewidth]{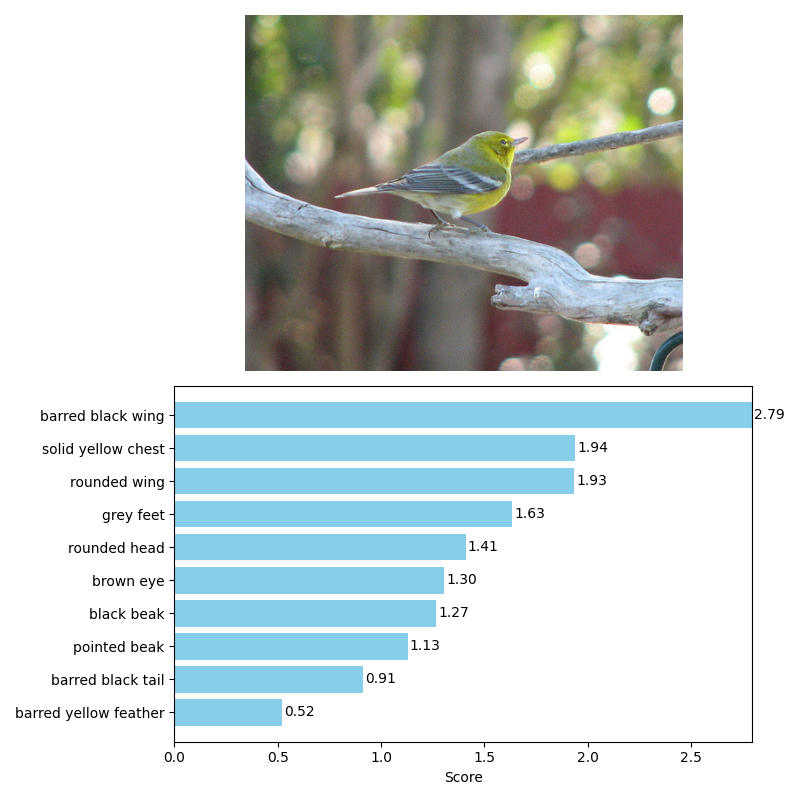}
\caption{Local Explanation Example of CUB-200-2011.}
\label{fig:local_cub_3}
\end{figure}

\begin{figure}[ht]
\centering
\includegraphics[width=\linewidth]{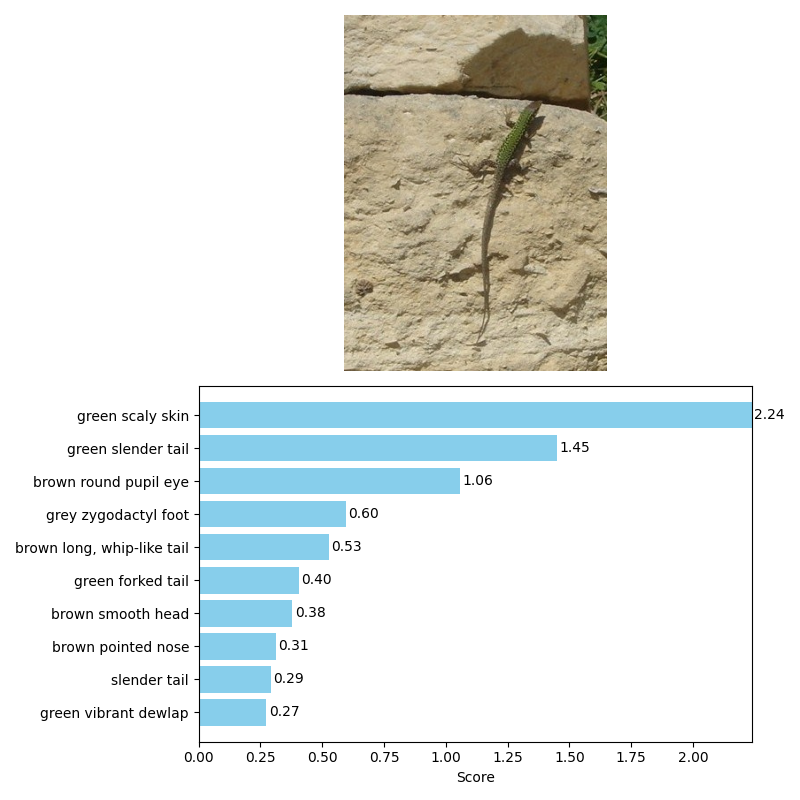}
\caption{Local Explanation Example of ImageNet-animal.}
\label{fig:local_in_1}
\end{figure}

\begin{figure}[ht]
\centering
\includegraphics[width=\linewidth]{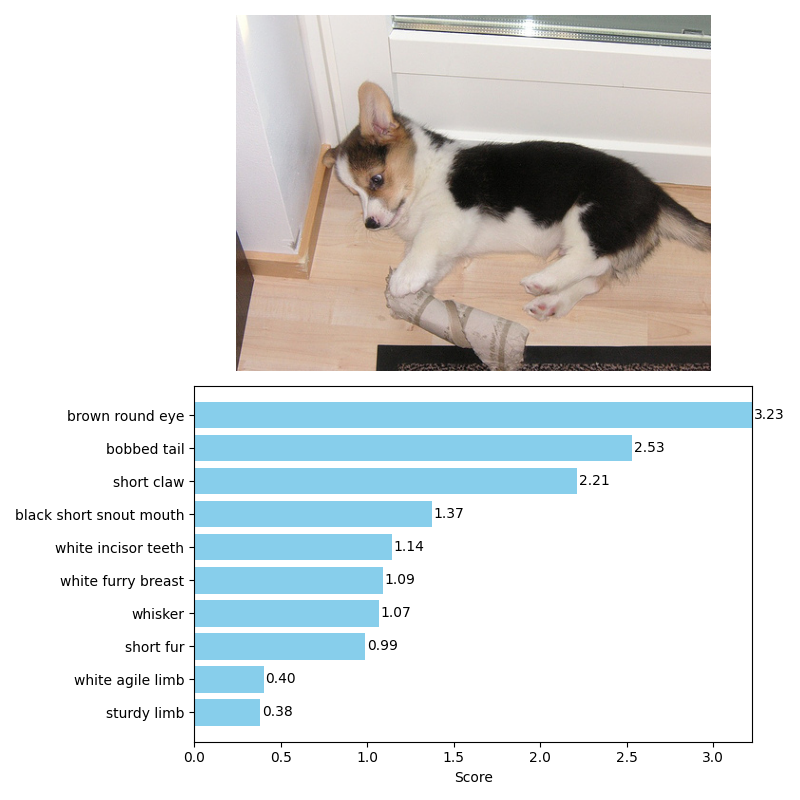}
\caption{Local Explanation Example of ImageNet-animal.}
\label{fig:local_in_2}
\end{figure}

\begin{figure}[ht]
\centering
\includegraphics[width=\linewidth]{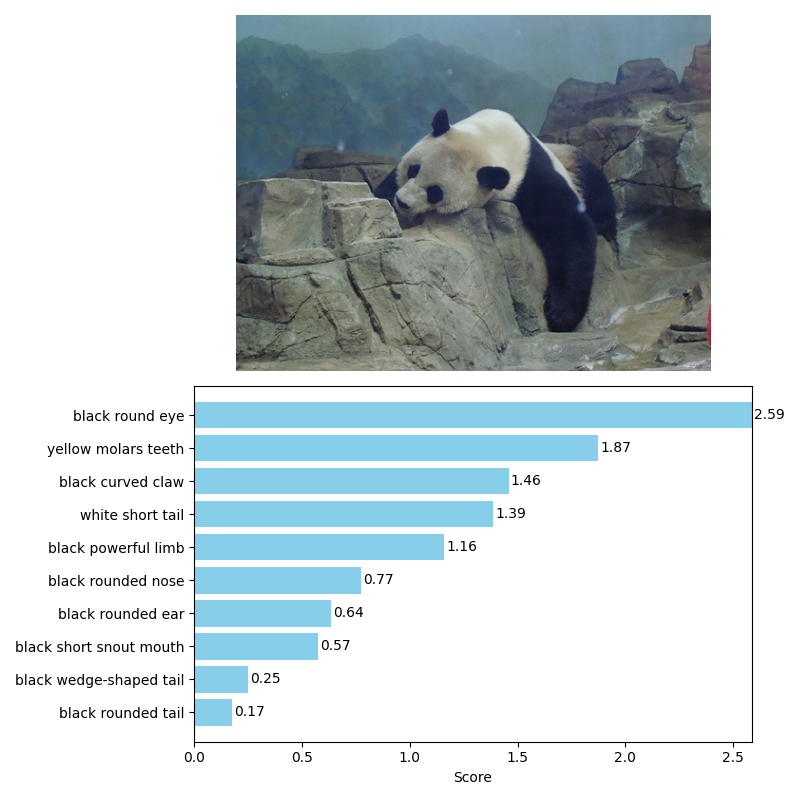}
\caption{Local Explanation Example of ImageNet-animal.}
\label{fig:local_in_3}
\end{figure}

\begin{figure}[ht]
\centering
\includegraphics[width=\linewidth]{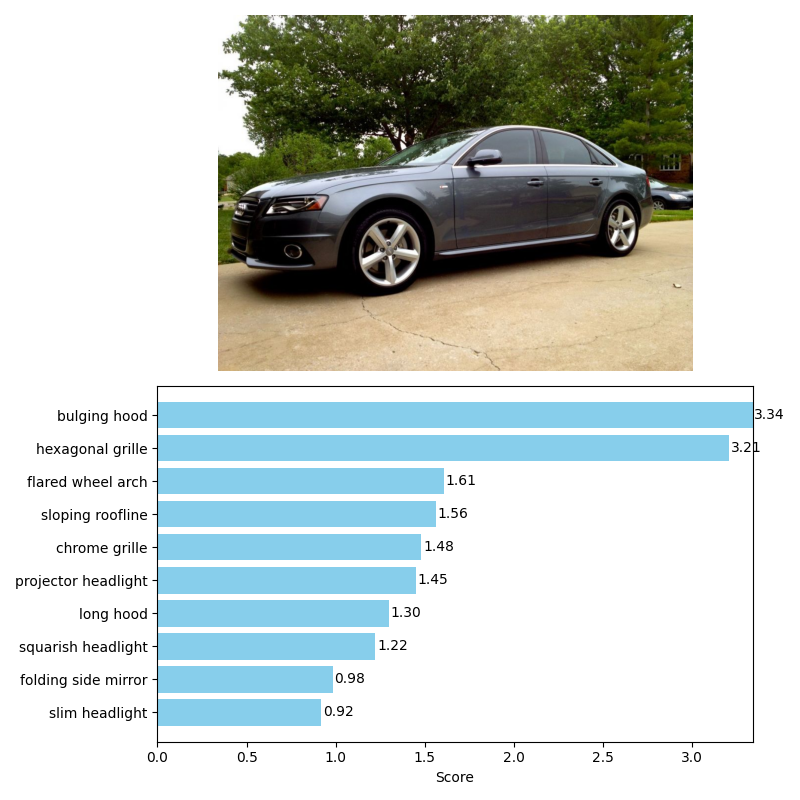}
\caption{Local Explanation Example of Stanford Cars.}
\label{fig:local_car_1}
\end{figure}

\begin{figure}[ht]
\centering
\includegraphics[width=\linewidth]{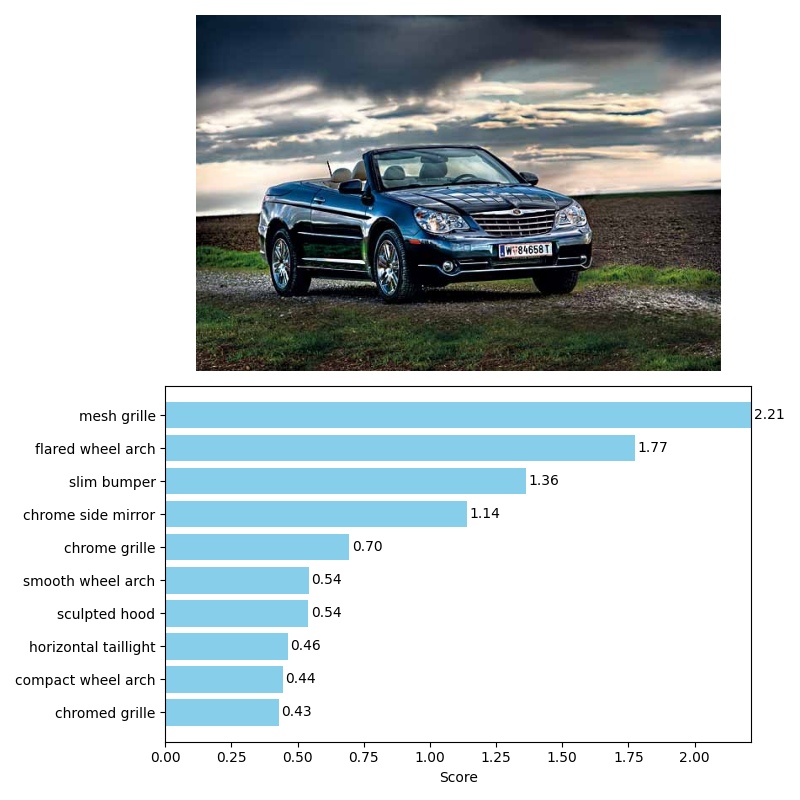}
\caption{Local Explanation Example of Stanford Cars.}
\label{fig:local_car_2}
\end{figure}

\begin{figure}[ht]
\centering
\includegraphics[width=\linewidth]{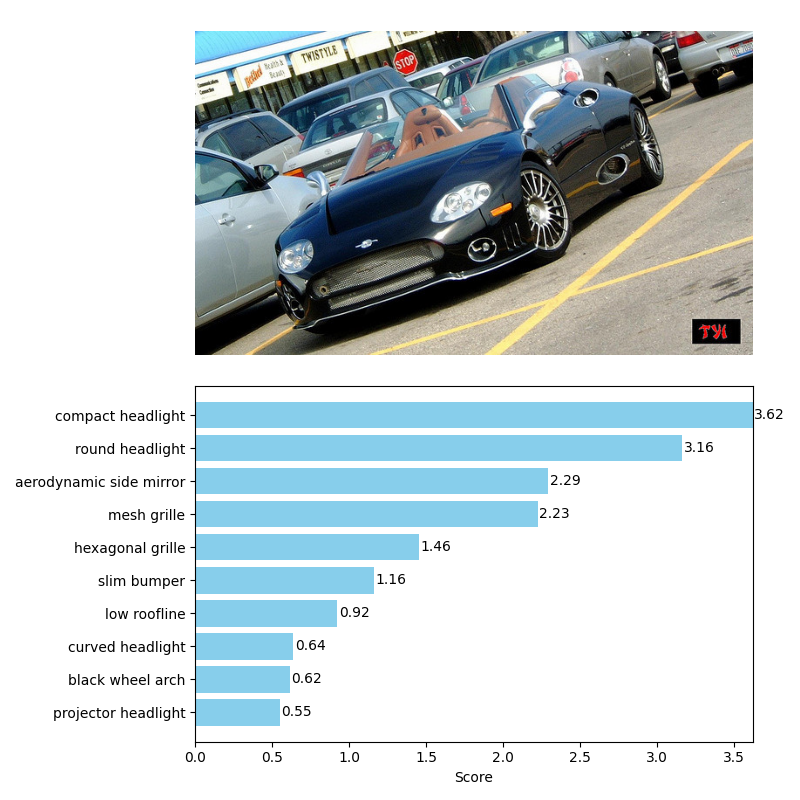}
\caption{Local Explanation Example of Stanford Cars.}
\label{fig:local_car_3}
\end{figure}

We present bar graphs illustrating how each concept contributes to the classification of individual images (i.e., local explanations). \Cref{fig:local_cub_1} to \Cref{fig:local_cub_3} show local explanation examples from CUB-200-2011, \Cref{fig:local_in_1} to \Cref{fig:local_in_3} from ImageNet-Animal, and \Cref{fig:local_car_1} to \Cref{fig:local_car_3} from StanfordCars. These visualizations indicate that \ours{} successfully identifies and highlights the key concepts present in each image.

\subsection{Discussions and Future Works}
Since obtaining annotations for concepts could be costly, recent works, including ours, utilize large pretrained models to extract them. Accurately measuring them would benefit from the rapid development of recent research on large language models and vision-language models. Another line of work has focused on the theoretical guarantee of identifying such factors \citep{scholkopf2021toward,rajendran2024causal,kim2025towards}.
On the other hand, having annotated concepts, recent works have focused on incorporating causality into concept bottleneck models, ensuring faithful explanations \citep{dominici2024counterfactual,hwang2025blackbox}.

We consider ensuring the robustness of localization in concept bottleneck models as a promising future direction. A potential way could be to incorporate data augmentations \citep{shorten2019survey,lee2022improving,hwang2022selecmix} or additional modalities such as segmentation maps \citep{kirillov2023segment} or sketches \citep{lee2023learning} during training.

%% file: main.bbl
\begin{thebibliography}{39}
\providecommand{\natexlab}[1]{#1}
\providecommand{\url}[1]{\texttt{#1}}
\expandafter\ifx\csname urlstyle\endcsname\relax
  \providecommand{\doi}[1]{doi: #1}\else
  \providecommand{\doi}{doi: \begingroup \urlstyle{rm}\Url}\fi

\bibitem[Achiam et~al.(2023)Achiam, Adler, Agarwal, Ahmad, Akkaya, Aleman, Almeida, Altenschmidt, Altman, Anadkat, et~al.]{gpt4}
Josh Achiam, Steven Adler, Sandhini Agarwal, Lama Ahmad, Ilge Akkaya, Florencia~Leoni Aleman, Diogo Almeida, Janko Altenschmidt, Sam Altman, Shyamal Anadkat, et~al.
\newblock Gpt-4 technical report.
\newblock \emph{arXiv preprint arXiv:2303.08774}, 2023.

\bibitem[Benou and Raviv(2025)]{salfcbm}
Itay Benou and Tammy~Riklin Raviv.
\newblock Show and tell: Visually explainable deep neural nets via spatially-aware concept bottleneck models.
\newblock In \emph{Proceedings of the Computer Vision and Pattern Recognition Conference}, pages 30063--30072, 2025.

\bibitem[Chen et~al.(2019)Chen, Li, Tao, Barnett, Rudin, and Su]{protopnet}
Chaofan Chen, Oscar Li, Daniel Tao, Alina Barnett, Cynthia Rudin, and Jonathan~K Su.
\newblock This looks like that: deep learning for interpretable image recognition.
\newblock \emph{Advances in neural information processing systems}, 32, 2019.

\bibitem[Deng et~al.(2009)Deng, Dong, Socher, Li, Li, and Fei-Fei]{imagenet}
Jia Deng, Wei Dong, Richard Socher, Li-Jia Li, Kai Li, and Li Fei-Fei.
\newblock Imagenet: A large-scale hierarchical image database.
\newblock In \emph{2009 IEEE Conference on Computer Vision and Pattern Recognition}, pages 248--255, 2009.

\bibitem[Dominici et~al.(2024)Dominici, Barbiero, Giannini, Gjoreski, Marra, and Langheinrich]{dominici2024counterfactual}
Gabriele Dominici, Pietro Barbiero, Francesco Giannini, Martin Gjoreski, Giuseppe Marra, and Marc Langheinrich.
\newblock Counterfactual concept bottleneck models.
\newblock \emph{arXiv preprint arXiv:2402.01408}, 2024.

\bibitem[Espinosa~Zarlenga et~al.(2022)Espinosa~Zarlenga, Barbiero, Ciravegna, Marra, Giannini, Diligenti, Shams, Precioso, Melacci, Weller, et~al.]{cem}
Mateo Espinosa~Zarlenga, Pietro Barbiero, Gabriele Ciravegna, Giuseppe Marra, Francesco Giannini, Michelangelo Diligenti, Zohreh Shams, Frederic Precioso, Stefano Melacci, Adrian Weller, et~al.
\newblock Concept embedding models: Beyond the accuracy-explainability trade-off.
\newblock \emph{Advances in Neural Information Processing Systems}, 35:\penalty0 21400--21413, 2022.

\bibitem[Groh et~al.(2021)Groh, Harris, Soenksen, Lau, Han, Kim, Koochek, and Badri]{fitz}
Matthew Groh, Caleb Harris, Luis Soenksen, Felix Lau, Rachel Han, Aerin Kim, Arash Koochek, and Omar Badri.
\newblock Evaluating deep neural networks trained on clinical images in dermatology with the fitzpatrick 17k dataset.
\newblock In \emph{Proceedings of the IEEE/CVF Conference on Computer Vision and Pattern Recognition}, pages 1820--1828, 2021.

\bibitem[He et~al.(2016)He, Zhang, Ren, and Sun]{resnet}
Kaiming He, Xiangyu Zhang, Shaoqing Ren, and Jian Sun.
\newblock Deep residual learning for image recognition.
\newblock In \emph{Proceedings of the IEEE conference on computer vision and pattern recognition}, pages 770--778, 2016.

\bibitem[Huang et~al.(2024)Huang, Song, Hu, Zhang, Wang, and Song]{protocbm}
Qihan Huang, Jie Song, Jingwen Hu, Haofei Zhang, Yong Wang, and Mingli Song.
\newblock On the concept trustworthiness in concept bottleneck models.
\newblock In \emph{Proceedings of the AAAI Conference on Artificial Intelligence}, pages 21161--21168, 2024.

\bibitem[Hwang et~al.(2022)Hwang, Lee, Kwak, Oh, Teney, Kim, and Zhang]{hwang2022selecmix}
Inwoo Hwang, Sangjun Lee, Yunhyeok Kwak, Seong~Joon Oh, Damien Teney, Jin-Hwa Kim, and Byoung-Tak Zhang.
\newblock Selecmix: Debiased learning by contradicting-pair sampling.
\newblock \emph{Advances in Neural Information Processing Systems}, 35:\penalty0 14345--14357, 2022.

\bibitem[Hwang et~al.(2025)Hwang, Pan, and Bareinboim]{hwang2025blackbox}
Inwoo Hwang, Yushu Pan, and Elias Bareinboim.
\newblock From black-box to causal-box: Towards building more interpretable models.
\newblock Technical Report R-127, Columbia CausalAI Laboratory, 2025.
\newblock Columbia CausalAI Laboratory, Technical Report (R-127).

\bibitem[Kim et~al.(2023)Kim, Jung, Park, Kim, and Yoon]{probcbm}
Eunji Kim, Dahuin Jung, Sangha Park, Siwon Kim, and Sungroh Yoon.
\newblock Probabilistic concept bottleneck models.
\newblock In \emph{International Conference on Machine Learning}, pages 16521--16540. PMLR, 2023.

\bibitem[Kim et~al.(2025)Kim, Nam, Hwang, and Lee]{kim2025towards}
Kwonho Kim, Heejeong Nam, Inwoo Hwang, and Sanghack Lee.
\newblock Towards causal representation learning with observable sources as auxiliaries.
\newblock In \emph{UAI 2025 Workshop on Causal Abstractions and Representations}, 2025.

\bibitem[Kirillov et~al.(2023)Kirillov, Mintun, Ravi, Mao, Rolland, Gustafson, Xiao, Whitehead, Berg, Lo, et~al.]{kirillov2023segment}
Alexander Kirillov, Eric Mintun, Nikhila Ravi, Hanzi Mao, Chloe Rolland, Laura Gustafson, Tete Xiao, Spencer Whitehead, Alexander~C Berg, Wan-Yen Lo, et~al.
\newblock Segment anything.
\newblock In \emph{Proceedings of the IEEE/CVF international conference on computer vision}, pages 4015--4026, 2023.

\bibitem[Koh et~al.(2020)Koh, Nguyen, Tang, Mussmann, Pierson, Kim, and Liang]{cbm}
Pang~Wei Koh, Thao Nguyen, Yew~Siang Tang, Stephen Mussmann, Emma Pierson, Been Kim, and Percy Liang.
\newblock Concept bottleneck models.
\newblock In \emph{International conference on machine learning}, pages 5338--5348. PMLR, 2020.

\bibitem[Krause et~al.(2013)Krause, Stark, Deng, and Fei-Fei]{cars}
Jonathan Krause, Michael Stark, Jia Deng, and Li Fei-Fei.
\newblock 3d object representations for fine-grained categorization.
\newblock In \emph{Proceedings of the IEEE international conference on computer vision workshops}, pages 554--561, 2013.

\bibitem[Lai et~al.(2024)Lai, Hu, Wang, Berti-Equille, and Wang]{faithful}
Songning Lai, Lijie Hu, Junxiao Wang, Laure Berti-Equille, and Di Wang.
\newblock Faithful vision-language interpretation via concept bottleneck models.
\newblock In \emph{The Twelfth International Conference on Learning Representations}, 2024.

\bibitem[Lee et~al.(2023)Lee, Hwang, Go, Choi, Kim, and Zhang]{lee2023learning}
Hyundo Lee, Inwoo Hwang, Hyunsung Go, Won-Seok Choi, Kibeom Kim, and Byoung-Tak Zhang.
\newblock Learning geometry-aware representations by sketching.
\newblock In \emph{Proceedings of the IEEE/CVF Conference on Computer Vision and Pattern Recognition}, pages 23315--23326, 2023.

\bibitem[Lee et~al.(2022)Lee, Hwang, Kang, and Zhang]{lee2022improving}
Sangjun Lee, Inwoo Hwang, Gi-Cheon Kang, and Byoung-Tak Zhang.
\newblock Improving robustness to texture bias via shape-focused augmentation.
\newblock In \emph{Proceedings of the ieee/cvf conference on computer vision and pattern recognition}, pages 4323--4331, 2022.

\bibitem[Li et~al.(2017)Li, Yang, Liu, Zhou, Wen, and Xu]{resnet_finetune}
Zhichao Li, Yi Yang, Xiao Liu, Feng Zhou, Shilei Wen, and Wei Xu.
\newblock Dynamic computational time for visual attention.
\newblock In \emph{Proceedings of the IEEE international conference on computer vision workshops}, pages 1199--1209, 2017.

\bibitem[Loshchilov and Hutter(2019)]{adamw}
Ilya Loshchilov and Frank Hutter.
\newblock Decoupled weight decay regularization.
\newblock In \emph{International Conference on Learning Representations}, 2019.

\bibitem[Ma et~al.(2023)Ma, Zhao, Chen, and Rudin]{protoconcepts}
Chiyu Ma, Brandon Zhao, Chaofan Chen, and Cynthia Rudin.
\newblock This looks like those: Illuminating prototypical concepts using multiple visualizations.
\newblock \emph{Advances in Neural Information Processing Systems}, 36:\penalty0 39212--39235, 2023.

\bibitem[Margeloiu et~al.(2021)Margeloiu, Ashman, Bhatt, Chen, Jamnik, and Weller]{intended}
Andrei Margeloiu, Matthew Ashman, Umang Bhatt, Yanzhi Chen, Mateja Jamnik, and Adrian Weller.
\newblock Do concept bottleneck models learn as intended?
\newblock \emph{arXiv preprint arXiv:2105.04289}, 2021.

\bibitem[Nauta et~al.(2023)Nauta, Schl\"otterer, van Keulen, and Seifert]{pipnet}
Meike Nauta, J\"org Schl\"otterer, Maurice van Keulen, and Christin Seifert.
\newblock Pip-net: Patch-based intuitive prototypes for interpretable image classification.
\newblock In \emph{Proceedings of the IEEE/CVF Conference on Computer Vision and Pattern Recognition (CVPR)}, pages 2744--2753, 2023.

\bibitem[Oikarinen et~al.(2023)Oikarinen, Das, Nguyen, and Weng]{labelfree}
Tuomas Oikarinen, Subhro Das, Lam~M. Nguyen, and Tsui-Wei Weng.
\newblock Label-free concept bottleneck models.
\newblock In \emph{The Eleventh International Conference on Learning Representations}, 2023.

\bibitem[Radford et~al.(2021)Radford, Kim, Hallacy, Ramesh, Goh, Agarwal, Sastry, Askell, Mishkin, Clark, et~al.]{clip}
Alec Radford, Jong~Wook Kim, Chris Hallacy, Aditya Ramesh, Gabriel Goh, Sandhini Agarwal, Girish Sastry, Amanda Askell, Pamela Mishkin, Jack Clark, et~al.
\newblock Learning transferable visual models from natural language supervision.
\newblock In \emph{International conference on machine learning}, pages 8748--8763, 2021.

\bibitem[Rajendran et~al.(2024)Rajendran, Buchholz, Aragam, Sch{\"o}lkopf, and Ravikumar]{rajendran2024causal}
Goutham Rajendran, Simon Buchholz, Bryon Aragam, Bernhard Sch{\"o}lkopf, and Pradeep Ravikumar.
\newblock From causal to concept-based representation learning.
\newblock \emph{Advances in Neural Information Processing Systems}, 37:\penalty0 101250--101296, 2024.

\bibitem[Raman et~al.(2024)Raman, Zarlenga, Heo, and Jamnik]{locality}
Naveen Raman, Mateo~Espinosa Zarlenga, Juyeon Heo, and Mateja Jamnik.
\newblock Do concept bottleneck models obey locality?, 2024.

\bibitem[Sch{\"o}lkopf et~al.(2021)Sch{\"o}lkopf, Locatello, Bauer, Ke, Kalchbrenner, Goyal, and Bengio]{scholkopf2021toward}
Bernhard Sch{\"o}lkopf, Francesco Locatello, Stefan Bauer, Nan~Rosemary Ke, Nal Kalchbrenner, Anirudh Goyal, and Yoshua Bengio.
\newblock Toward causal representation learning.
\newblock \emph{Proceedings of the IEEE}, 109\penalty0 (5):\penalty0 612--634, 2021.

\bibitem[Selvaraju et~al.(2017)Selvaraju, Cogswell, Das, Vedantam, Parikh, and Batra]{gradcam}
Ramprasaath~R Selvaraju, Michael Cogswell, Abhishek Das, Ramakrishna Vedantam, Devi Parikh, and Dhruv Batra.
\newblock Grad-cam: Visual explanations from deep networks via gradient-based localization.
\newblock In \emph{Proceedings of the IEEE international conference on computer vision}, pages 618--626, 2017.

\bibitem[Shang et~al.(2024)Shang, Zhou, Zhang, Ni, Yang, and Wang]{rescbm}
Chenming Shang, Shiji Zhou, Hengyuan Zhang, Xinzhe Ni, Yujiu Yang, and Yuwang Wang.
\newblock Incremental residual concept bottleneck models.
\newblock In \emph{Proceedings of the IEEE/CVF Conference on Computer Vision and Pattern Recognition}, pages 11030--11040, 2024.

\bibitem[Sheth and Kahou(2023)]{coopcbm}
Ivaxi Sheth and Samira~Ebrahimi Kahou.
\newblock Auxiliary losses for learning generalizable concept-based models.
\newblock In \emph{Thirty-seventh Conference on Neural Information Processing Systems}, 2023.

\bibitem[Shorten and Khoshgoftaar(2019)]{shorten2019survey}
Connor Shorten and Taghi~M Khoshgoftaar.
\newblock A survey on image data augmentation for deep learning.
\newblock \emph{Journal of big data}, 6\penalty0 (1):\penalty0 1--48, 2019.

\bibitem[Srivastava et~al.(2024)Srivastava, Yan, and Weng]{vlgcbm}
Divyansh Srivastava, Ge Yan, and Tsui-Wei Weng.
\newblock Vlg-cbm: Training concept bottleneck models with vision-language guidance.
\newblock \emph{arXiv preprint arXiv:2408.01432}, 2024.

\bibitem[Wah et~al.(2011)Wah, Branson, Welinder, Perona, and Belongie]{cub}
C. Wah, S. Branson, P. Welinder, P. Perona, and S. Belongie.
\newblock Caltech-ucsd birds-200-2011.
\newblock Technical Report CNS-TR-2011-001, California Institute of Technology, 2011.

\bibitem[Xu et~al.(2024)Xu, Qin, Mi, Wang, and Li]{ecbm}
Xinyue Xu, Yi Qin, Lu Mi, Hao Wang, and Xiaomeng Li.
\newblock Energy-based concept bottleneck models: Unifying prediction, concept intervention, and probabilistic interpretations.
\newblock In \emph{The Twelfth International Conference on Learning Representations}, 2024.

\bibitem[Yan et~al.(2023)Yan, Wang, Zhong, Dong, He, Lu, Wang, Shang, and McAuley]{concise}
An Yan, Yu Wang, Yiwu Zhong, Chengyu Dong, Zexue He, Yujie Lu, William~Yang Wang, Jingbo Shang, and Julian McAuley.
\newblock Learning concise and descriptive attributes for visual recognition.
\newblock In \emph{Proceedings of the IEEE/CVF International Conference on Computer Vision}, pages 3090--3100, 2023.

\bibitem[Yang et~al.(2023)Yang, Panagopoulou, Zhou, Jin, Callison-Burch, and Yatskar]{labo}
Yue Yang, Artemis Panagopoulou, Shenghao Zhou, Daniel Jin, Chris Callison-Burch, and Mark Yatskar.
\newblock Language in a bottle: Language model guided concept bottlenecks for interpretable image classification.
\newblock In \emph{Proceedings of the IEEE/CVF Conference on Computer Vision and Pattern Recognition}, pages 19187--19197, 2023.

\bibitem[Yuksekgonul et~al.(2023)Yuksekgonul, Wang, and Zou]{pcbm}
Mert Yuksekgonul, Maggie Wang, and James Zou.
\newblock Post-hoc concept bottleneck models.
\newblock In \emph{The Eleventh International Conference on Learning Representations}, 2023.

\end{thebibliography}
